\documentclass[10pt]{article} % For LaTeX2e
\usepackage[accepted]{rlc}
\usepackage{amssymb}            % Defines common symbols like \mathbb R
\usepackage{mathtools}          % Extends amsmath, providing common math tools
\usepackage{mathrsfs}           % Enables \mathscr, which can work in cases that \mathcal does not
\mathtoolsset{showonlyrefs}     % Only number equations that are referenced (optional)
\usepackage{graphicx}           % For including images
\usepackage{subcaption}         % Allows for the use of subfigures and subcaptions
\usepackage[space]{grffile}     % For spaces in image names
\usepackage{url}                % For displaying urls
\usepackage{makecell}
\usepackage{xspace}
\newcommand{\middlemoe}{$\textrm{Middle}$\xspace}
\newcommand{\finalmoe}{$\textrm{Final}$\xspace}
\newcommand{\allmoe}{$\textrm{All}$\xspace}
\newcommand{\bigmoe}{$\textrm{Big}$\xspace}

\newcommand\blfootnote[1]{%
  \begingroup
  \renewcommand\thefootnote{}\footnote{#1}%
  \addtocounter{footnote}{-1}%
  \endgroup
}

\newcommand{\sidebysideequations}[3][0.5]{%
  \begin{equation*}
  \begin{minipage}{#1\displaywidth}%
  \begin{equation}\vphantom{\def\label##1{}#3}#2\end{equation}
  \end{minipage}
  \begin{minipage}{\dimexpr\displaywidth-#1\displaywidth}
  \begin{equation}\vphantom{\def\label##1{}#2}#3\end{equation}
  \end{minipage}
  \end{equation*}
}
\usepackage[capitalise]{cleveref}

\title{Mixture of Experts in a Mixture of RL settings}

\author{Timon Willi\(^{*1,2,6}\), Johan Obando-Ceron\(^{*3,4,6}\), Jakob Foerster\(^{1,2}\), Karolina Dziugaite\(^{5,6}\),\\\textbf{Pablo Samuel Castro\(^{3,4,6}\)
}\\\\
Foerster Lab for AI Research\(^{1}\)\\
University of Oxford\(^{2}\)\\
Mila - Québec AI Institute\(^{3}\) \\
Universit\'e de Montr\'eal\(^{4}\) \\
McGill University\(^{5}\) \\
Google DeepMind\(^{6}\)  \\
}
\begin{document}

\maketitle
\blfootnote{*Authors contributed equally. Correspondence to \texttt{[timon.willi, jobando0730]@gmail.com},\texttt{psc@google.com}}

\begin{abstract}
Mixtures of Experts (MoEs) have gained prominence in (self-)supervised learning due to their enhanced inference efficiency, adaptability to distributed training, and modularity. Previous research has illustrated that MoEs can significantly boost Deep Reinforcement Learning (DRL) performance by expanding the network's parameter count while reducing dormant neurons\footnote{\textit{Dormant neurons:} neurons that have become practically inactive
through low activations.}, thereby enhancing the model's learning capacity and ability to deal with non-stationarity.
In this work, we shed more light on MoEs' ability to deal with non-stationarity and investigate MoEs in DRL settings with ``amplified'' non-stationarity via multi-task training, providing further evidence that MoEs improve learning capacity. In contrast to previous work, our multi-task results allow us to better understand the underlying causes for the beneficial effect of MoE in DRL training, the impact of the various MoE components, and insights into how best to incorporate them in actor-critic-based DRL networks. Finally, we also confirm results from previous work.
\end{abstract}

\section{Introduction}
\label{sec:introduction}

Deep Reinforcement Learning (RL), which integrates reinforcement learning algorithms with deep neural networks, has demonstrated remarkable success in enabling agents to achieve complex tasks beyond human capabilities in domains ranging from video games to strategic board games and beyond \citep{mnih2015humanlevel, berner2019dota, vinyals2019grandmaster, fawzi2022discovering, Bellemare2020AutonomousNO}. Despite the pivotal role of deep networks in these advanced RL applications, their learning dynamics within RL contexts still need to be fully understood. Recent research has uncovered unexpected behaviours and phenomena associated with the use of deep networks in RL, which often diverge from those observed in traditional supervised learning environments \citep{ostrovski2021tandem,kumar2021implicit, lyle2022understanding,graesser2022state,nikishin22primacy,sokar2023dormant,ceron2023small}.

Transformers \citep{vaswani2017attention}, adapters \citep{houlsby19parameter}, and Mixture of Experts \citep[MoEs;][]{shazeer2017outrageously}, are crucial for the scalability of supervised learning models, particularly within the domains of natural language processing and computer vision. MoEs stand out by facilitating the scaling of networks to encompass trillions of parameters, a feat made possible through their modular design that seamlessly integrates with distributed computing techniques~\citep{fedus2022switch}. Moreover, MoEs introduce a form of structured sparsity into the network architecture, a characteristic associated with enhancements in network performance through various studies on network sparsity \citep{evci2020rigl,gale2019state,jin2022pruning}. Finally, there is growing evidence in the supervised learning literature that MoEs specialise on different problem characteristics in {\em multi-task} settings \citep{gupta2022sparsely}.
%\psc{Find citations for this claim}. 
These settings are inherently non-stationary and may benefit from the modularity and sparsity induced by MoE-based architectures.

Recently, \citet{obando2024mixtures} demonstrated that MoEs unlock scaling in DRL networks for single-task settings. However, they did so under a specific setting where only the penultimate layer was replaced by an MoE module. Their analyses suggest that incorporating MoEs makes networks less susceptible to loss of plasticity, as evidenced by measurements including the fraction of dormant neurons. \citet{sokar2023dormant}, in exploring the phenomenon of dormant neurons in DRL, provided strong evidence that their growth is due mainly to the non-stationary nature of RL training.

In this work, we set out to better understand {\em how MoEs help training under non-stationarity} and {\em which aspects of MoEs yield these results}. To do so, we ``amplify'' the non-stationarity of DRL training by investigating settings where multiple tasks are learned concurrently by the same agent. Specifically, we investigate the incorporation of a variety of MoE architectures in Multi-Task Reinforcement Learning (MTRL) and Continual Reinforcement Learning (CRL) settings. Our results demonstrate that the induced sparsity of expert modules is critical to mitigating plasticity loss under amplified non-stationarity and highlight the difficulty and importance of properly training the router. While focusing on the MTRL and CRL settings, some insights below apply to more traditional single-task settings.

\section{Background}
\subsection{Reinforcement Learning}
A Markov Decision Process \citep[MDP;][]{bellman1957markovian, puterman1990markov, sutton2018reinforcement} 
%consists of 
is defined by
a tuple $\mathcal{M}=\langle\mathcal{S}, \mathcal{A}, \mathcal{P}, r, \rho, \gamma\rangle$, where $\mathcal{S}$ denotes the set of all possible states, $\mathcal{A}$ denotes the set of possible actions, $\mathcal{P}: \mathcal{S} \times \mathcal{A} \rightarrow \mathcal{S}$ 
is the state transition probability kernel, 
%distribution of transitioning to 
%specifying the probability that state $s'$ from state $s$ after taking action $a$. The function 
$r: \mathcal{S} \times \mathcal{A} \rightarrow \mathbb{R}$ is the reward function,
%received after taking an action $a$ in state $s$, 
$\rho$ denotes the initial state distribution, 
and $\gamma$ (where $0 < \gamma \leq 1$) is the discount factor that determines the present value of future rewards. 
In Reinforcement Learning (RL), a policy $\pi$ assigns to each state $s$ a probability distribution $\pi(s)$ over actions in $\mathcal{A}$. The objective in RL is to devise a policy $\pi$ that maximises the expected sum of discounted rewards $J(\pi)=\mathbb{E}_\pi\left[\sum_{t=0}^{\infty} \gamma^t r\left(s_t, a_t\right)\right]$. The policy parameterised by $\theta$ is denoted as $\pi_\theta(a_t \mid s_t)$.
%is parameterized by $\theta$.
The parameter $\theta$ is chosen via optimisation maximising $J(\theta)$, thereby achieving the highest possible cumulative reward.

In \textbf{Multi-Task Reinforcement Learning} (MTRL), an agent engages with a variety of tasks $\tau$ from a set $\mathcal{T}$, with each task $\tau$ constituting a distinct Markov Decision Process (MDP) denoted by $\mathcal{M}^\tau = \langle \mathcal{S}^\tau, \mathcal{A}^\tau, \mathcal{P}^\tau, r^\tau, \rho^\tau, \gamma^\tau \rangle$. The objective in MTRL is to devise a unified policy $\pi$ that optimises the average expected cumulative discounted return across all tasks, expressed as $J(\theta) = \frac{1}{|\mathcal{T}|}\sum_\tau J_\tau(\theta)$. %The tasks may vary across one or several MDP components, with task-specific information being provided either through a task identifier, like a one-hot vector, or via metadata, such as a task description, following the approach by \citet{sodhani2021multi}. 
In our work, at each training step a single agent trains synchronously on multiple tasks. MTRL is effective at measuring an agent's capabilities at devising control policies from a highly varying set of inputs and environment dynamics.

\textbf{Continual RL} \citep{abbas2023loss} is a variant of MTRL where the agent trains on one task for an extended period before switching to a new task \citep{khetarpal2022towards}; once all tasks have been trained on once, the agent once again trains on all the environments in the same order. This setting enables measuring an agent's ability to learn new tasks while retaining previously learned policies.

As a concrete example, imagine we have two MDPs ($\mathcal{M}^1$, $\mathcal{M}^2$). MTRL would train on both $(\mathcal{M}^1, \mathcal{M}^2)$ at each step, whereas CRL would train on $\mathcal{M}^1$ for an extended number of steps, then $\mathcal{M}^2$, and so on: $\mathcal{M}^1\rightarrow\mathcal{M}^2\rightarrow\mathcal{M}^1\rightarrow\mathcal{M}^2$.

\begin{figure}[!t]
    \centering
    \includegraphics[width=\textwidth]{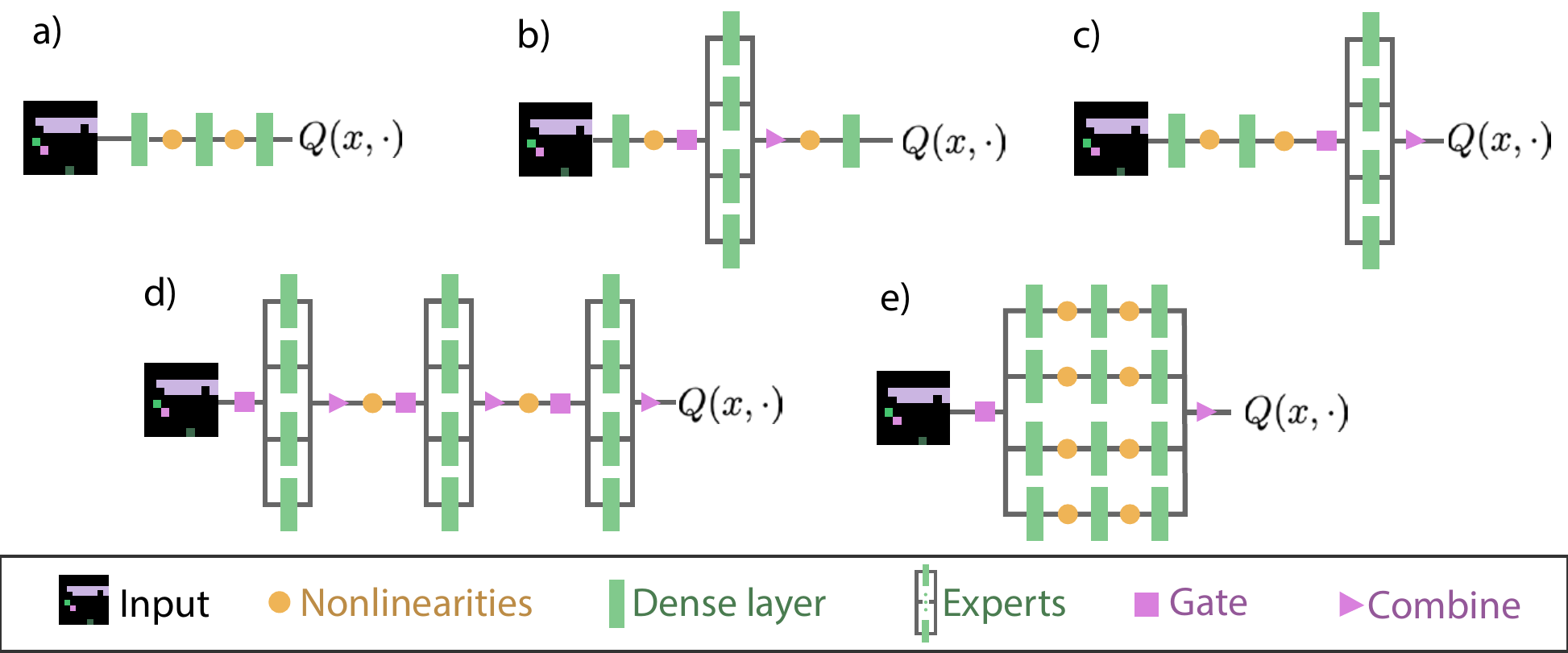}
    \caption{\textbf{Architectures considered: (a)} Baseline architecture; \textbf{(b)} \middlemoe,  used by \citet{obando2024mixtures}; \textbf{(c)} \finalmoe, where an MoE module replaces the final layer; \textbf{(d)} \allmoe, where all layers are replaced with an MoE module; \textbf{(e)} \bigmoe, with a single MoE module where an expert comprises the full original network.}
    \label{fig:moe_architectures}
    \vspace{-0.5em}
\end{figure}

\subsection{MoEs}
\label{sec:moeBackground}
Mixtures of Experts (MoEs) emerged as a cornerstone in designing Large Language Models (LLMs), integrating a collection of $n$ ``expert'' sub-networks. 
A gating mechanism, known as the router and usually learned during training, manages the experts by directing each incoming token to $k$ selected experts \citep{shazeer2017outrageously}. Typically, $k$ is less than the total count of experts (in our case, $k=1$). This sparsity is key for enhancing inference speed and facilitating distributed computing, making it a pivotal factor in training LLMs. In transformer architectures, MoE units substitute all dense feedforward layers \citep{vaswani2017attention}. The impressive empirical performance of MoEs has sparked significant research interest \citep{shazeer2017outrageously,Lewis2021BASELS,fedus2022switch,zhou2022mixture,puigcerver2023sparse,lepikhin2020gshard,zoph2022stmoe,gale2023megablocks}.

The strict routing of tokens to specific experts, known as hard assignments, presents several issues, including training instability, token loss, and obstacles in expanding the number of experts \citep{fedus2022switch,puigcerver2023sparse}. To mitigate these issues, \citet{puigcerver2023sparse} proposed the concept of \textbf{Soft MoE}, which utilises a soft, fully differentiable method for allocating tokens to experts, thereby circumventing the limitations associated with router-based hard assignments. This soft assignment method calculates a blend of weights for each token across the experts and aggregates their outputs accordingly. 
Adopting the terminology of \citet{puigcerver2023sparse}, consider input tokens represented by $ \mathbf{X} \in \mathbb{R}^{m \times d} $, with $ m $ indicating the count of $ d $-dimensional tokens. A Soft MoE layer processes these tokens through $ n $ experts, each defined as $ \{f_i: \mathbb{R}^d \rightarrow \mathbb{R}^d\}_{1: n} $. Every expert is associated with $ p $ slots for both input and output, each slot characterised by a $ d $-dimensional vector of parameters. These parameters are collectively denoted as $ \boldsymbol{\Phi} \in \mathbb{R}^{d \times(n \cdot p)} $.

The input slots $\tilde{\mathbf{X}} \in \mathbb{R}^{(n \cdot p) \times d}$ represent a weighted average of all tokens, given by $\tilde{\mathbf{X}}=\mathbf{D}^{\top} \mathbf{X}$, where $\mathbf{D}$ is commonly known as the dispatch weights. The outputs from the experts are expressed as $ \tilde{\mathbf{Y}}_i = f_{\lfloor i / p\rfloor}(\tilde{\mathbf{X}}_i) $. For the Soft MoE layer, the overall output $ \mathbf{Y} $ results from merging $ \tilde{\mathbf{Y}} $ with the combined weights $ \mathbf{C} $, described by $ \mathbf{Y} = \mathbf{C} \tilde{\mathbf{Y}} $. $\mathbf{D}$ and $\mathbf{C}$ are represented by the following expressions:

\sidebysideequations[0.6]{
            \mathbf{D}_{i j}=\frac{\exp \left((\mathbf{X} \boldsymbol{\Phi})_{i j}\right)}{\sum_{i^{\prime}=1}^m \exp \left((\mathbf{X} \boldsymbol{\Phi})_{i^{\prime} j}\right)}, \notag
            }{\mathbf{C}_{i j}=\frac{\exp \left((\mathbf{X} \boldsymbol{\Phi})_{i j}\right)}{\sum_{j^{\prime}=1}^{n \cdot p} \exp \left((\mathbf{X} \boldsymbol{\Phi})_{i j^{\prime}}\right)} \notag.
            }
The findings from \citet{puigcerver2023sparse} indicate that Soft MoE provides an improved balance between accuracy and computational expense relative to alternative MoE approaches.

\section{Mixtures of Experts in a mixture of RL settings}
\label{sec:moesHelp}
Although shifting targets (due to bootstrapping) and dynamic data collection (from the agent's policy) already render single-task RL a non-stationary problem, MTRL and CRL take this non-stationarity to an extreme by changing the environments during training. A critical difference between the two is that in MTRL, at every step, the agent interacts and learns from each environment (e.g. regular task-switching). In contrast, in CRL tasks are switched very infrequently. Thus, both settings provide complementary perspectives when investigating the efficacy of MoEs under high levels of non-stationarity.

\subsection{Experimental setup} 
As we investigate many settings in many scenarios, we wanted to maximise the number of runs per setting to ensure statistical robustness, while keeping the computational expense at bay. For this reason, we chose to run our experiments with the PureJaxRL codebase\footnote{PureJaxRL code available at: \href{https://github.com/luchris429/purejaxrl}{https://github.com/luchris429/purejaxrl}} \citep{lu2022model, lu2022discovered, lu2023adversarial}, which is a high-performance and parallelisable library including an implementation of Proximal Policy Optimisation \citep[PPO]{schulman2017proximal}. Since \citet{obando2024mixtures} focused on value-based methods, our use of PPO provides complementary insights and results. We rely on the Gymnax suite \citep{gymnax2022github} to implement optimised versions of MinAtar environments \citep{young19minatar}, which have been shown to provide insights comparable to the full ALE suite \citep{ceron2021revisiting}. The hyper-parameters used are provided in \autoref{app:hyperparams} and were adapted from \citet{jesson2023relu} (we deviate in the network size due to computational constraints). For all experiments, we evaluate on three environments: SpaceInvaders (SI), Breakout (BO), and Asterix (Ast). The input observations from Asterix differ substantially from SpaceInvaders and Breakout, whereas the latter are similar in observation and action space. This environment selection allows us to investigate whether MDP similarity encourages sharing representations between experts.

%#mtrl_hardcoded2
%netsModel2
\begin{figure*}[!t]
    \centering
    \includegraphics[width=\textwidth]{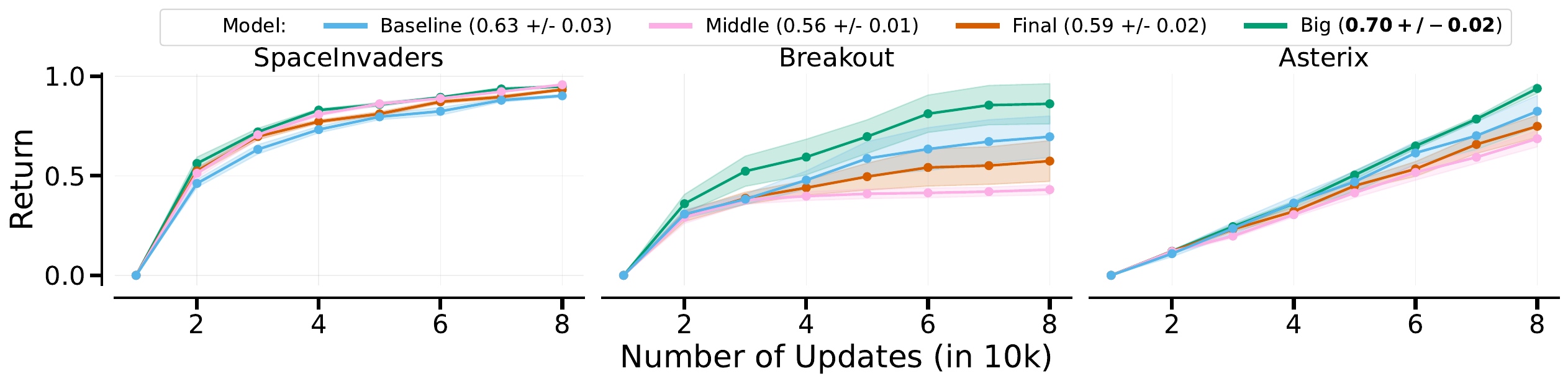}
    \vspace{3mm}
    \includegraphics[width=\textwidth]{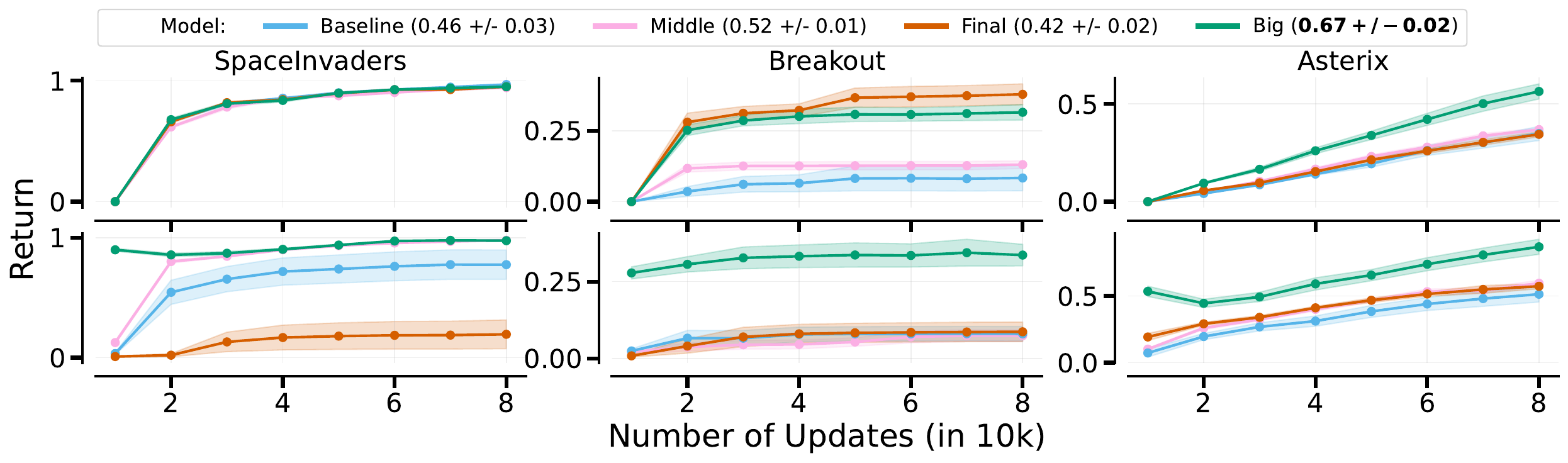}
    \caption{\textbf{Measuring the impact of MoE architectures} with hardcoded routing in MTRL (top). and CRL (bottom). In each legend, the numbers in parentheses indicate the average performance of each approach over all games. \bigmoe{} outperforms all other methods.}
    %\vspace{-3mm}}
    \label{fig:moe_architecture_impact}
    %\vspace{-0.5em}
\end{figure*}

\begin{figure*}[!t]
    \centering
    \includegraphics[width=\textwidth]{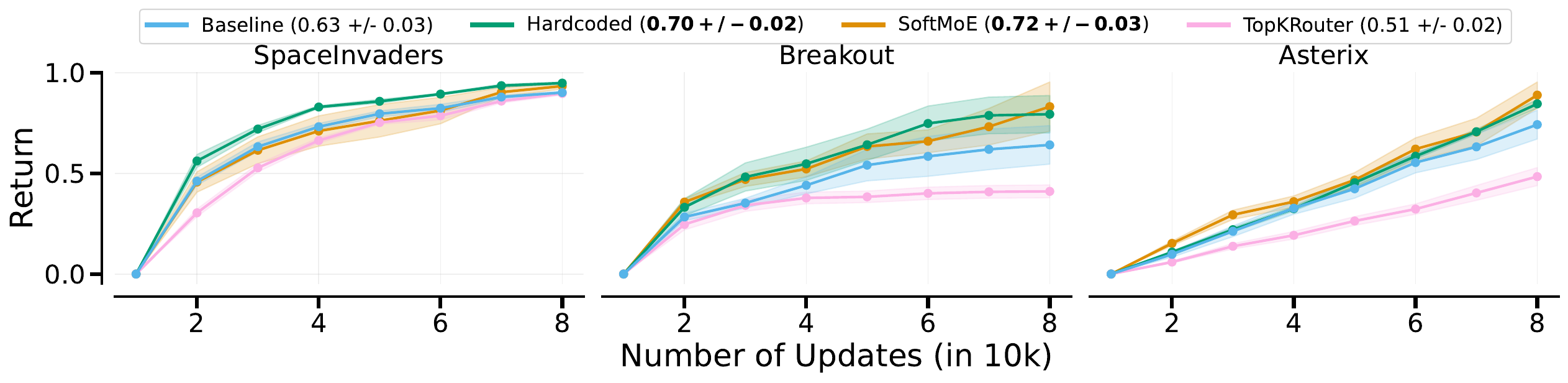}
    %Continual_MTRL2.pdf}
    \vspace{3mm}
    \includegraphics[width=\textwidth]{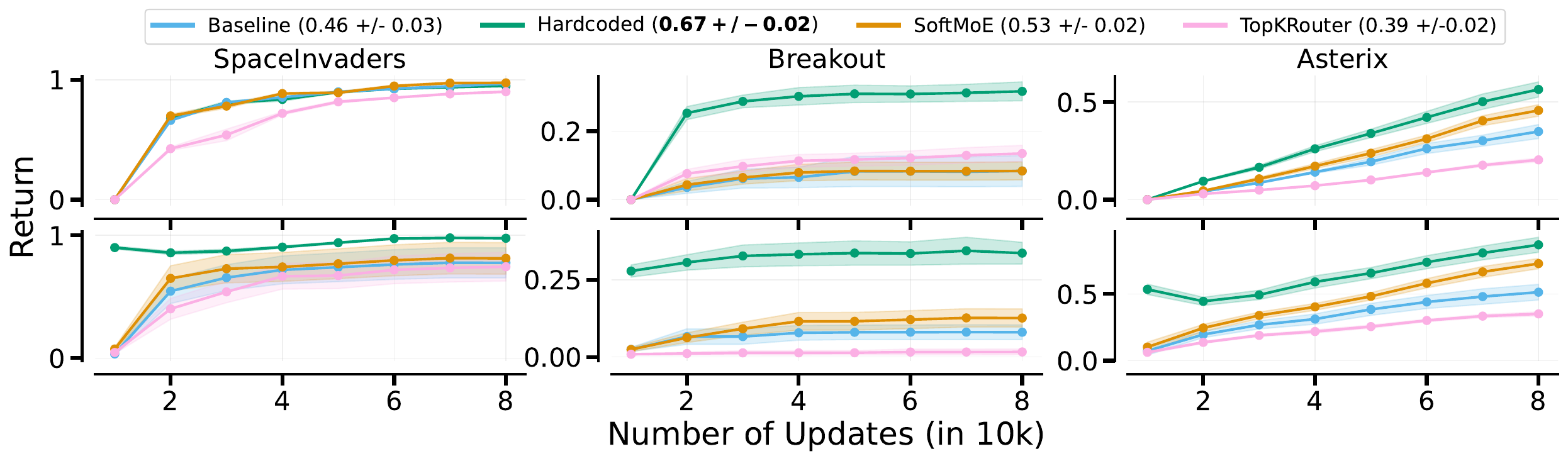}
    \caption{\textbf{Measuring the impact of routing} with \bigmoe{} architecture using different routing approaches under the MTRL (top row) and CRL (bottom row) settings. In each legend, the numbers in parentheses indicate the average performance of each approach across all games. SoftMoE and Hardcoded work best in MTRL, and Hardcoded works best in CRL, though SoftMoE still outperforms the baseline.\vspace{-3mm}}
    \label{fig:crl_top}
    %\vspace{-0.5em}
\end{figure*}

For \textbf{MTRL} we train simultaneously on SI, BO, and Ast; in practice, the PPO agent performs one update step per environment in sequence.
For \textbf{CRL}, we train the agent on a fixed sequence of MDPs \citep{abbas2023loss} for $1e7$ environment steps ($\sim$ 80k update steps), specifically SI $\rightarrow$ BO $\rightarrow$ Ast $\rightarrow$ SI $\rightarrow$ BO $\rightarrow$ Ast. We present further analysis with different task orders in \cref{sec:analysis}.

\citet{obando2024mixtures} propose replacing the penultimate layer with an MoE module and sharing the other layers across the network. We term this variant \middlemoe. We also evaluate a variant called \finalmoe, where the MoE module replaces the last layer. We also propose two new architectures: \allmoe, where MoE modules replace all three layers, and \bigmoe, where the network contains a single MoE module and each expert consists of a full network (see \autoref{fig:moe_architectures}). Since we are dealing with three distinct environments, all versions of MoEs have three experts. In all cases, we are using per-sample tokenization: one token -- the state -- per forward pass \citep{obando2024mixtures}. All other hyper-parameters are reported in \autoref{app:hyperparams}.

We use a hardcoded routing strategy for many of our experiments to isolate the impact of {\em routing} versus {\em expert architecture}. This routing strategy will assign one expert for each task and route inputs accordingly. For the \bigmoe architecture, this effectively trains a separate network for each task and serves as a useful baseline. For all our results, we report the mean, averaged over $10$ independent seeds, with shaded areas representing standard error. In most figures, we also present the average normalised performance across all tasks (in parentheses in the legend), where normalisation scores were taken from \citet{jesson2023relu}. Our experiments were run on a single Tesla P$100$ or A$100$ GPU, each taking $10$ minutes. In total, we ran $870$ distinct settings over $10$ seeds each and are reported in \cref{sec:impact_hard,sec:impact_learned,sec:analysis,app:crl1,app:mtrl1,app:ac_ablation,app:order_ablation,app:mtrl_more_results,app:crl_more_results,app:single_env_more_results}.

\subsection{The impact of MoE architectures}
\label{sec:impact_hard}
To evaluate the impact of the choice of MoE architectures (\autoref{fig:moe_architectures}), we make use of the hardcoded router, which avoids potentially confounding factors due to also learning a routing strategy. In \autoref{fig:moe_architecture_impact}, it is evident that the \bigmoe{} architecture performs best as expected since it is essentially a separate network per task. Still, it is promising to observe that all architectures outperform the baseline in the CRL setting, with \allmoe{} being the strongest performer. Surprisingly, in the MTRL setting, only \bigmoe{} outperforms the baseline. We hypothesise that \allmoe{} struggles due to suboptimal hyperparameters, as it was not computationally feasible to run a hyperparameter search over all possible settings. For \middlemoe{} and \finalmoe{}, it is possible that gradient interference \citep{lyle23understanding} is complicating the learning process since there is parameter sharing outside of the MoE modules.

\subsection{The impact of learned routers}
\label{sec:impact_learned}
The \bigmoe{} architecture provides a direct way to evaluate the impact of routing, as gating and combining are only done before and after the original network parameters, respectively. In \autoref{fig:crl_top}, we present the learning curves for \bigmoe{} architecture with varying routing strategies under CRL and MTRL.

In MTRL, we see little difference between the hardcoded router and SoftMoE. This is surprising since the hardcoded router trains separate networks for each task (performing as well as the baseline trained on each environment individually). This suggests that the gating used by SoftMoE is effective in situations where tasks are frequently changed. The rigidity of TopK routing appears to make it difficult for it to learn proper routing strategies, resulting in deteriorated performance.

In CRL, the hardcoded router performs best and retains previously learned policies (as evidenced by the second time the tasks are run). While SoftMoE ultimately outperforms the baseline in each task, it struggles in retaining previously learned policies; it is worth noting that the second time training on Ast, although its starting performance is essentially at zero, its final performance is higher than the first time training through, suggesting some policy retention (bottom right of \cref{fig:crl_top}).

A major learning challenge in the CRL setting is that no signal is provided to the network when the environment changes. Thus, a natural question is whether learned routers can effectively use task information.
To investigate this, in \autoref{fig:gradient_analysis}, we added the task ID as an input to the router (top row) and observed the surprising result that including task ID slightly hurts performance for \bigmoe{}-SoftMoE. Examining the gradient similarity from one update to the next (bottom left panel of \autoref{fig:gradient_analysis}), it becomes evident that task-switching induces a discontinuity in the gradients used for learning. Interestingly, including this gradient similarity information as part of the input to the router does not hurt performance, but it does not improve either (bottom right).

In summary, our results suggest that SoftMoE routing is effective at dealing with high levels of non-stationarity, provided that discontinuous changes in environment dynamics (such as those arising from task switching) occur with relative frequency.

\section{Extra Analyses}
\label{sec:analysis}

\begin{figure}[!t]
    \centering
    \includegraphics[width=\textwidth]{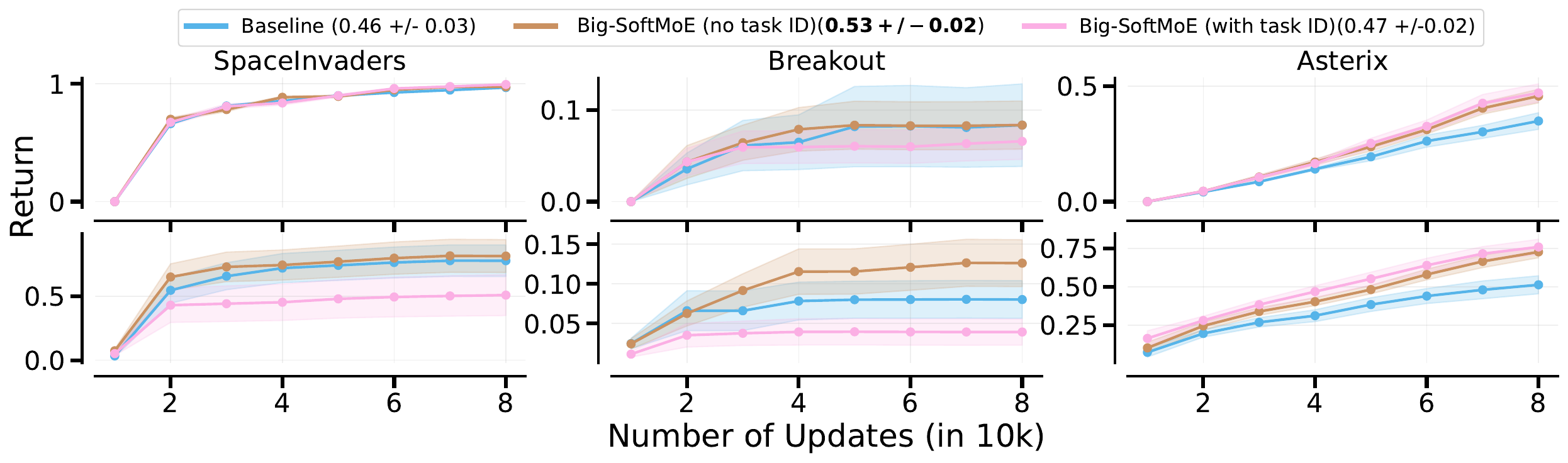}
    \vspace{3mm}
     \includegraphics[width=0.3\textwidth]{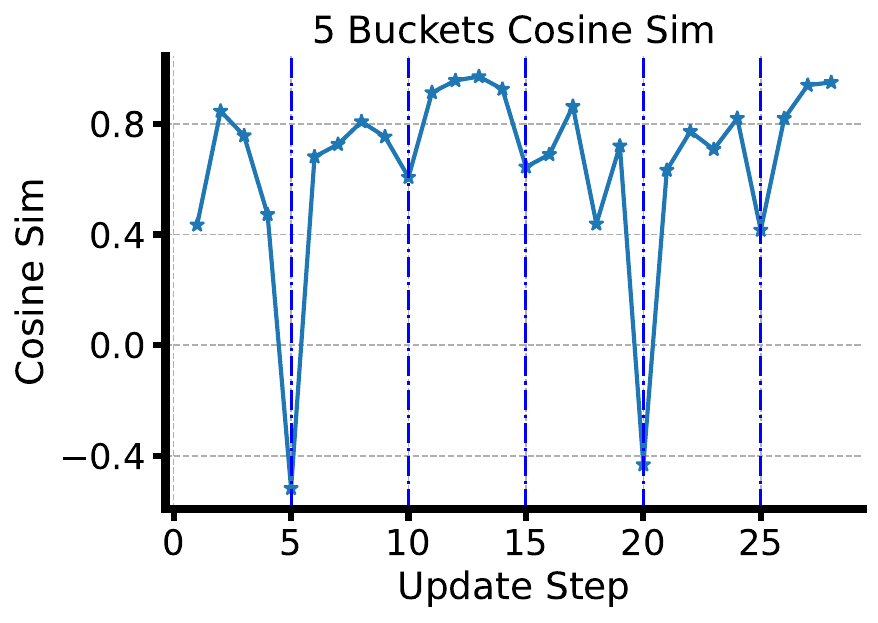}%
   \vline height 95pt depth 0 pt width 1.2 pt
   \includegraphics[width=0.7\textwidth]{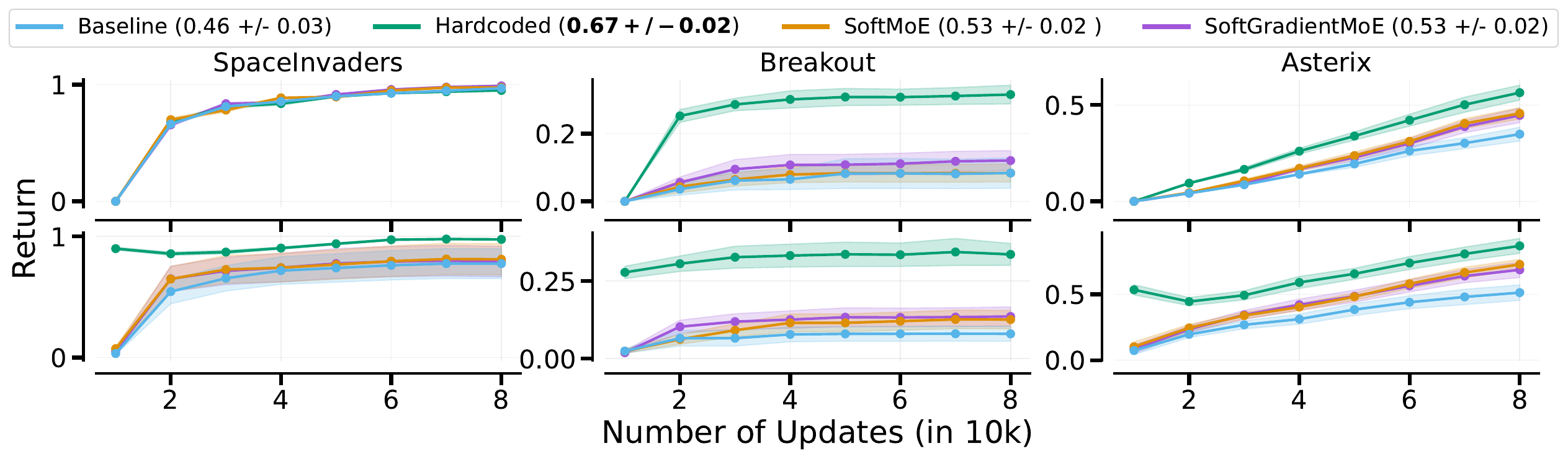}%
   \vspace{3mm}
   \caption{\textbf{Top:} Adding the task ID as an input to the router hurts performance for Big-SoftMoE. \textbf{Bottom left:} Sequential gradient similarity calculated throughout training, where dashed vertical lines represent when tasks switch. \textbf{Bottom right:} Adding gradient information as an input to the router does not improve performance.\vspace{-1mm}}
   \label{fig:gradient_analysis}
   %\vspace{-1em}
   \vspace{-0.5em}
\end{figure}
In the previous section, we provided empirical evidence suggesting that MoEs can improve DRL agents' performance in various non-stationary training regimes. Next, we conduct additional analyses to uncover the underlying causes of MoEs' benefits.

\begin{figure}[!t]
    \centering
    % \includegraphics[width=0.53\textwidth]{figures/dormantNeurons_MTRL2.pdf}%
    % \vline height 60pt depth 0 pt width 1.2 pt
    \includegraphics[width=\textwidth]{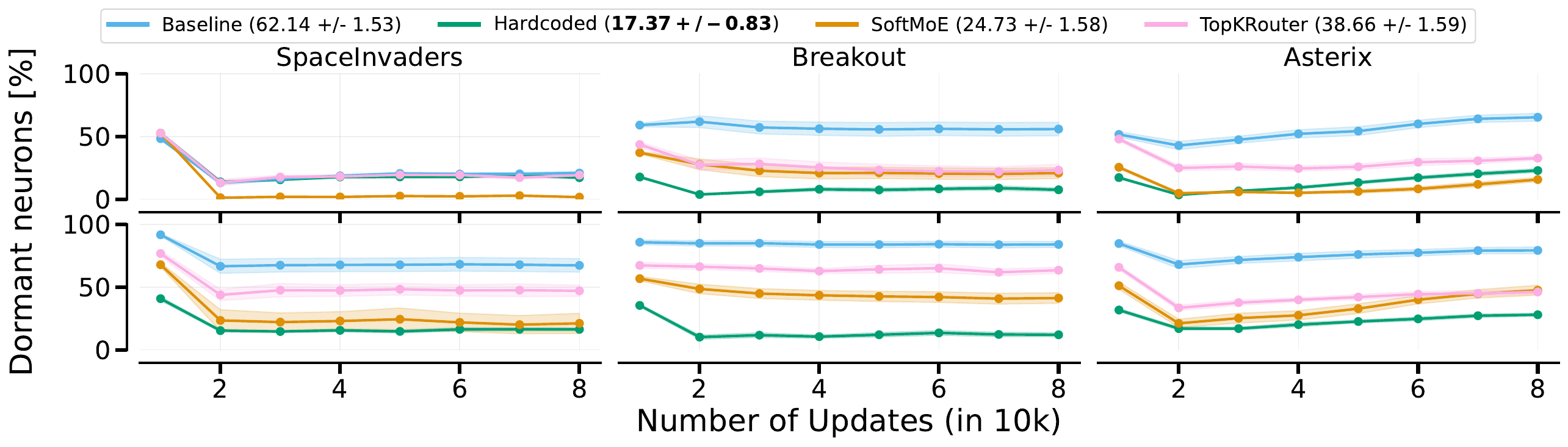}
    \vspace{3mm}
  \includegraphics[width=\textwidth]{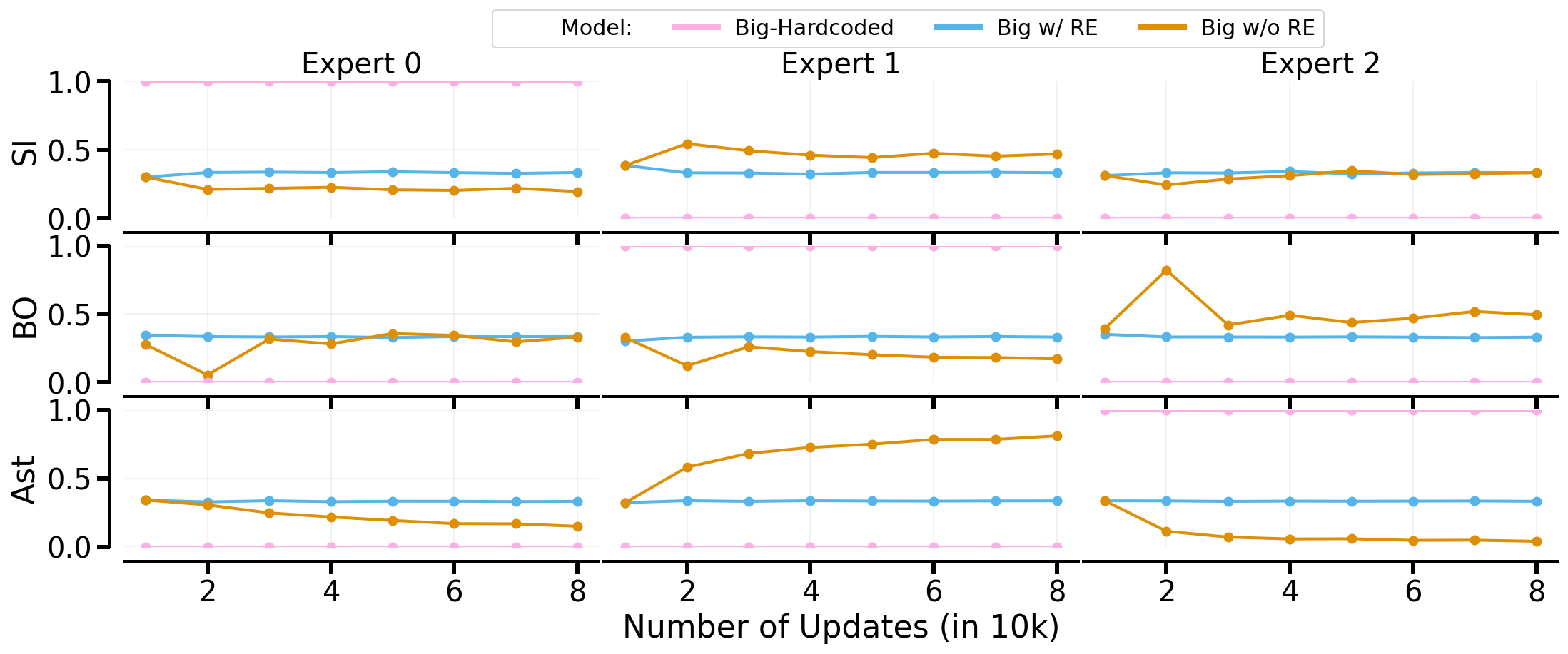}%
  \caption{\textbf{Top:} presents the ratio of dormant neurons for CRL under different routing approaches using \bigmoe{}. The numbers in the legend represent average dormant neuron fractions across all games. MoE variants have lower dormant neurons than the baseline. \textbf{Bottom:} Regularising the entropy of the router makes the expert selection more uniform. Without regularisation, there is more specialisation. This shows one seed, as different seeds might choose different experts. See \autoref{sec:analysis} for more details.}
  \label{fig:dormant_and_experts}
  %\vspace{-1em}
  \vspace{-0.5em}
\end{figure}

\textbf{Impact on network plasticity.}
We measure the fraction of dormant neurons \citep{sokar2023dormant} during training as a proxy for network plasticity. As Figs.~\ref{fig:dormant_and_experts} (top), \ref{fig:crl_dormant}, and \ref{fig:mtrl_dormant}  demonstrate, all MoE variants reduce the fraction of dormant neurons, suggesting MoEs help with maintaining network plasticity, consistent with the findings of \citet{obando2024mixtures}.

\textbf{Expert specialisation.} In Figs.~\ref{fig:dormant_and_experts} (bottom), \ref{fig:crl_big_expert}, and  \ref{fig:mtrl_expert}  we measure the probabilities assigned to each expert during training; what these values indicate is the likelihood that inputs will be routed to each respective expert; observe that the hardcoded router has maximal specialisation, where each expert is assigned to one task. We can also observe that both \bigmoe{}-SoftMoE and \allmoe-SoftMoE variants tend to specialise in all layers. 

In supervised learning settings, it is common to use load-balancing losses to {\em avoid} this type of specialisation to maximise expert usage. We explored this idea by adding entropy regularisation during training and observed that, while we do see a decrease in expert specialisation (c.f. Figs.~\ref{fig:gradient_analysis} (bottom), \ref{fig:crl_big_expert}, and \ref{fig:mtrl_expert}), this does not affect performance in any meaningful way
(c.f. \cref{fig:mtrl_re,tab:mtrl_re}).
%Given the limited success of specialisation, we hypothesise that encouraging more random routing via entropy regularisation of the router might help.  
%Unfortunately, this approach also falls short, failing to produce a meaningful improvement in performance for \bigmoe-SoftMoE and \allmoe-SoftMoE. See \autoref{fig:mtrl_re} and \autoref{tab:mtrl_re}.
%The expert selection probabilities are shown in \cref{fig:gradient_analysis,fig:mtrl_expert,fig:crl_big_expert}.

\begin{figure*}[!t]
    \centering
    \includegraphics[width=\textwidth]{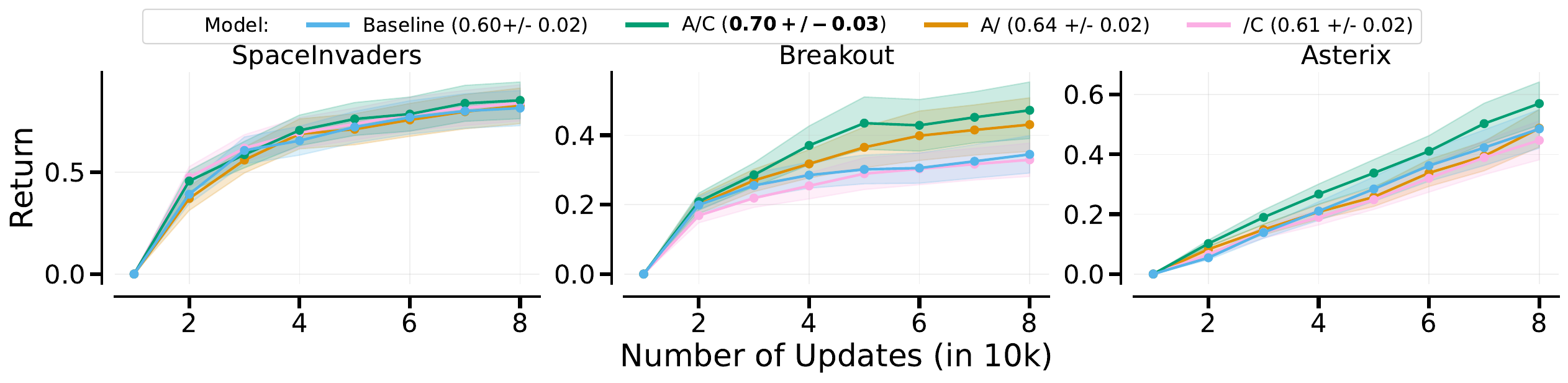}
    \includegraphics[width=\textwidth]{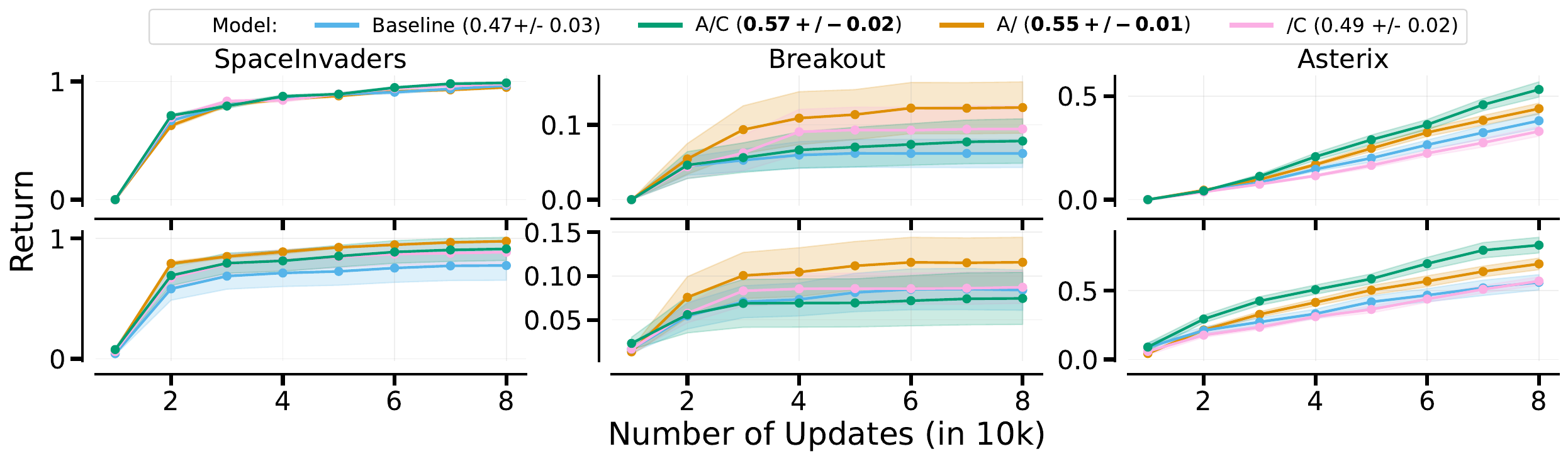}
    %ContinualRL_AC2.pdf}
    %\vspace{-0.3mm}
    \caption{\textbf{Comparing the impact of MoE architectures on actor and critic networks} with the hardcoded router, under the MTRL (top row) and CRL (bottom row) settings. In each legend, the numbers in parentheses indicate the average performance of each approach across all games. It is best to use MoEs on both networks. However, the results suggest that MoEs have a greater impact when used on the actor network than on the critic. See \autoref{sec:analysis} for more details.} %\vspace{-3mm}}
    \label{fig:crl_ac}
    %\vspace{-0.5em}
\end{figure*}

\textbf{Impact on actor and critic networks.}
\citet{obando2024mixtures} focused on value-based methods (where a single network serves as critic and actor), so using an actor-critic method like PPO provides a novel, complementary perspective. By default, we use MoE modules on the actor and critic networks, but in \cref{fig:crl_ac,fig:crl_ac2,fig:mtrl_ac,tab:crl_ac,tab:mtrl_ac}, we show that, in the two settings, it is best to use MoEs on both networks. However, the results suggest that MoEs have a greater impact on the actor than on the critic network. The fact that actor networks seem to benefit more from MoEs than critic networks is aligned with the findings of \citet{graesser2022state}, where they found that actor networks could handle much higher levels of sparsity than critics without any degradation in performance.

\textbf{Order of environments.}
To investigate the impact of environment ordering, we train using the ordering Ast $\rightarrow$ BO $\rightarrow$ SI to compare with the orderings we have used thus far; we present results for MTRL and CRL in \cref{fig:crl_order,fig:mtrl_order}, respectively. While conclusions do not change in MTRL, changing the order of environments affects CRL performance significantly (excluding the hardcoded router). We observe two interesting changes: (i) training BO after Ast (as opposed to after SI) causes all methods (excluding hardcoded) to collapse, and (ii) when training on Ast last, none of the agents were able to retain the learned policy (c.f. \cref{fig:crl_top}), whereas when training on Ast first there is some policy retention (as seen on the bottom left of \cref{fig:crl_order}). As mentioned previously, Ast differs substantially from the other two environments, so our findings in \cref{fig:crl_order} suggest that the agents have overfit the input distribution of Ast, hindering its ability to adapt to the other environments but allowing the retention of the policy learned on Ast.

\textbf{Single Environment.}
% Some evidence that BigMoE with SoftMoE helps performance
Despite the clear improvements from CRL and MTRL, there are no significant performance improvements across all games in the single environment setting. However, \bigmoe improves over the baseline in Asterix while performing worse in Breakout, as shown in \autoref{fig:single_env_top} and \autoref{tab:single_env}, suggesting that MoEs might be beneficial in specific types of environments. Adding gradient information did not affect performance (see \autoref{fig:single_env_vanilla}).

\section{Related Work}
% We want to investigate if MoEs help with some of the many shortcomings of current Deep RL techniques, such as capacity loss and catastrophic forgetting. Next, we will provide an overview of the settings and methods that related work proposed to address said problems.
%
% Shortcomings of RL
%\textbf{Parameter Efficiency in Deep RL.} 
Parameter underutilisation is a roadblock to parameter efficiency in deep Reinforcement Learning (RL).
The latter was highlighted by \citet{sokar2023dormant} in the form of dormant neurons. %Parameter underutilisation was also observed in terms of achievable sparsity.
\citet{arnob2021single} demonstrate that in offline RL, up to 95\% of network parameters can be pruned at initialisation without impacting performance. 
Further, several studies have shown that periodic network weight resets enhance performance \citep{igl2020transient, dohare2021continual, nikishin2022primacy, doro2022sample, sokar2023dormant, schwarzer23bbf} and that RL networks maintain performance when trained with a high degree of sparsity \citep{tan2022rlx2, sokar2022dynamic, graesser2022state, obando2024deep}. 
These findings underscore the need for methods that more effectively leverage network parameters in RL training. Our work explores the use of Mixture of Experts (MoEs) for actor-critic methods, demonstrating significant reductions in dormant neurons across various tasks and network architectures.

Mixtures of Experts (MoEs) revolutionised large-scale language/vision models primarily due to their modular design, which supports distributed training and enhances parameter efficiency during inference \citep{lepikhin2020gshard, fedus2022switch, yang2019condconv, wang2020deep, abbas2020biased, pavlitskaya2020using}. MoEs show benefits in transfer and multi-task learning scenarios, e.g., by assigning experts to specific sub-problems \citep{puigcerver2023sparse, chen2023mod, ye2023taskexpert}, or by improving the statistical performance of routers \citep{hazimeh2021dselect}.

MoEs have been studied in DRL \citep{ren2021probabilistic, hendawy2024multitask, akrour2021continuous} but based on a previous definition of MoE \citep{jacobs1991adaptive}, closely related to ensembling, and not the more recent interpretation of MoEs in LLMs. Ensembles are often used to represent the policy \citep{anschel2016deep, lan2020maxmin, agarwal2020optimistic, peer2021ensemble, chen2021randomized, an2021uncertainty, wu2021aggressive, liang2022reducing} or to predict model dynamics \citep{shyam2019model, chua2018deep, kurutach2018model}. Most closely related to ensembling is our \bigmoe{} architecture, where each expert is a full model. \citet{fan2023learnable} could be interpreted as using multiple meta-controllers as routers for \bigmoe{} and ensembling the resulting policy. In contrast to our work, they do not investigate different MoE architectures and rely on population-based training.

Two recent works have explored using MoEs (as used in LLMs) in DRL: the work of \citet{obando2024mixtures} has already been referenced extensively above, as our work builds on their findings. More recently, \citet{farebrother2024stop} argued that classification losses yield stabler learning dynamics than regression losses, which also applies to using MoEs.

\section{Conclusion}
Our work provides additional evidence of the effectiveness of MoEs in improving the training of DRL agents. Using MTRL and CRL grants us a novel perspective on evaluating and analysing MoEs under ``extreme'' non-stationarity. Consistent with the findings of \citet{obando2024mixtures}, DRL is most performant using SoftMoE, whereas it struggles with hard TopK routing.

Our use of the hardcoded router served as a useful baseline for our analyses and demonstrates much room for improvement in training DRL agents in multi-task settings. Indeed, in CRL, only mild policy retention was observed in Ast, and the retention amount was dependent on the order in which the environments were trained. An exciting avenue for future work is thus investigating what {\em task curricula} would lead to best agent performance and policy retention. As mentioned previously, the observations in Ast differ substantially from those of BO and SI (which are similar to each other); the fact that we only observed policy retention in Ast thus begs the question of whether the agent is over-fitting to the anomalous input distribution of Ast, at the expense of being able to generalise to the other environments.

Expert specialisation and whether load-balancing is desirable are also interesting questions for future research. The findings from the supervised learning community in this respect may not naturally carry over to DRL settings, largely due to training's inherent non-stationarity. Finally, MoEs could be investigated in multi-agent settings, where experts represent different agents in cooperative \citep{ellis2024smacv2} or general-sum settings \citep{lu2022model, willi2022cola}, where vectorised environments are widely available \citep{khan2023scaling, rutherford2023jaxmarl}.

% Switching for continual RL is unsolved, Hardcoded Router shows promising results
%The challenge of task switching in Continual Reinforcement Learning (CRL) remains largely unresolved. However, the Hardcoded Router and learned variants have shown promising results in some settings, indicating a potential pathway towards more effective strategies for managing task transitions within this context. This approach stands out for its ability to provide a structured mechanism for switching between tasks, addressing one of CRL's most persistent challenges.

% SoftMoE do not provide performance benefits for MTRL out of the gate
%In the Multi-Task Reinforcement Learning (MTRL) setting, applying \bigmoe{}-SoftMoE improves performance over the baseline and the hardcoded variant. This finding prompts a deeper exploration into the conditions under which SoftMoE effectively leverages task-similarity within MTRL frameworks.

% TopKRouter struggles across the board.
%Conversely, the TopKRouter faces challenges consistently across various tasks. Its struggle to perform effectively underscores the complexities involved in designing routers that can adeptly manage the distribution of tasks among experts.

%There are many avenues left to explore task switching for CRL, building on the gradient analysis of this work. In another direction, 

\subsubsection*{Acknowledgements}

The authors would like to thank Gheorghe Comanici, Gopeshh Subbaraj, Doina Precup, Hugo Larochelle, and the rest of the Google DeepMind Montreal team for valuable discussions during the preparation of this work.  Gheorghe Comanici deserves a special mention for providing us valuable feed-back on an early draft of the paper. We thank the anonymous reviewers for their valuable help in improving our manuscript. We would also like to thank the Python community \cite{van1995python, 4160250} for developing tools that enabled this work, including NumPy \cite{harris2020array}, Matplotlib \cite{hunter2007matplotlib}, Jupyter \cite{2016ppap}, Pandas \cite{McKinney2013Python} and JAX \cite{bradbury2018jax}.

\subsubsection*{Broader Impact Statement}
\label{sec:broaderImpact}
This paper introduces research aimed at pushing the boundaries of Machine Learning, with a particular focus on reinforcement learning. Our work holds various potential societal implications, although we refrain from singling out specific ones for emphasis in this context. 

% \label{sec:ack}
% Use unnumbered third level headings for the acknowledgments. All acknowledgments, including those to funding agencies, go at the end of the paper. Only add this information once your submission is accepted and deanonymized. 
% \input{oldStructure}

\bibliography{main}
\bibliographystyle{rlc}

\clearpage
\appendix
\section{Continual RL}
\label{app:crl1}
\begin{figure*}[hbt!]
    \centering
    \includegraphics[width=\textwidth]{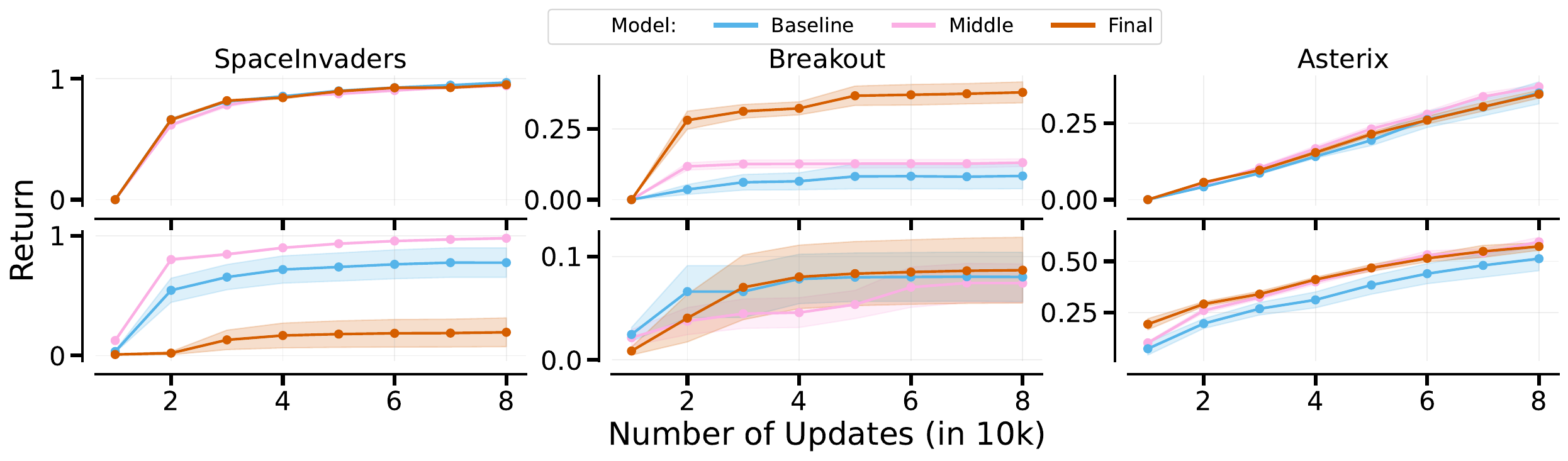}
    \caption{\textbf{CRL: Variants with shared parameters across task} do not necessarily improve performance over the baseline. \middlemoe performs better than the baseline, whereas \finalmoe does not.}
    \label{fig:crl_shared_params}
\end{figure*}

\begin{table}[hbt!]
\centering
\begin{tabular}{|l|c|c|c|}
\hline
Game & Baseline & Final-Hardcoded & Middle-Hardcoded \\
\hline
SI & $0.97 \pm 0.01$ & $0.95 \pm 0.01$ & $0.94 \pm 0.01$ \\
BO & $0.08 \pm 0.05$ & $0.38 \pm 0.04$ & $0.13 \pm 0.01$ \\
Ast & $0.35 \pm 0.04$ & $0.34 \pm 0.01$ & $0.37 \pm 0.01$ \\
SI-2 & $0.78 \pm 0.12$ & $0.19 \pm 0.12$ & $0.98 \pm 0.01$ \\
BO-2 & $0.08 \pm 0.02$ & $0.09 \pm 0.03$ & $0.07 \pm 0.02$ \\
Ast-2 & $0.51 \pm 0.06$ & $0.57 \pm 0.02$ & $0.60 \pm 0.02$ \\
\hline
Total & $0.46 \pm 0.03$ & $0.42 \pm 0.02$ & $\textbf{0.52} \pm \textbf{0.01}$ \\
\hline
\end{tabular}
\caption{\textbf{CRL: Variants with shared parameters across task} do not necessarily improve performance over the baseline. \middlemoe performs better than the baseline, whereas \finalmoe does not.}
\label{tab:crl_shared_params}
\end{table}

\begin{figure*}[hbt!]
    \centering
    \includegraphics[width=\textwidth]{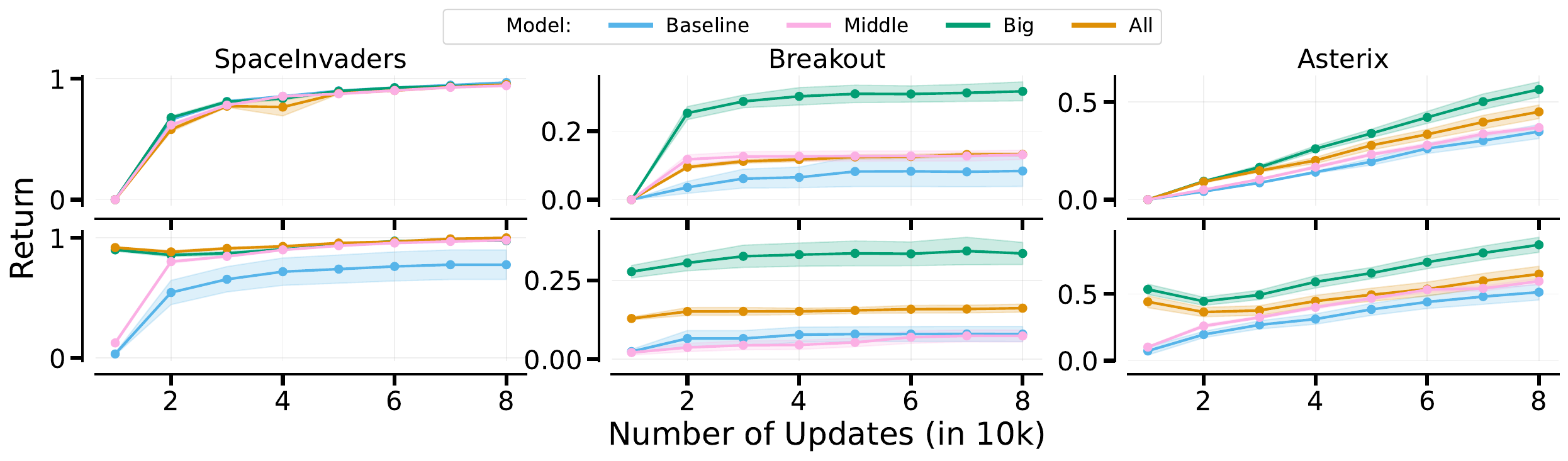}
    \caption{\textbf{CRL: Isolated Params: Variants with isolated parameters across task} improve performance over the baseline. \bigmoe-Hardcoded works the best.}
    \label{fig:crl_isolated_params}
\end{figure*}

\begin{table}[hbt!]
\centering
\begin{tabular}{|l|c|c|c|c|}
\hline
Game & Baseline & All-Hardcoded & Big-Hardcoded & Middle-Hardcoded \\
\hline
Game & Baseline & Final-Hardcoded & Middle-Hardcoded & Additional-Column \\
\hline
SI & $0.97 \pm 0.01$ & $0.95 \pm 0.01$ & $0.95 \pm 0.00$ & $0.94 \pm 0.01$ \\
BO & $0.08 \pm 0.05$ & $0.13 \pm 0.00$ & $0.32 \pm 0.03$ & $0.13 \pm 0.01$ \\
Ast & $0.35 \pm 0.04$ & $0.45 \pm 0.03$ & $0.56 \pm 0.04$ & $0.37 \pm 0.01$ \\
SI-2 & $0.78 \pm 0.12$ & $1.00 \pm 0.00$ & $0.98 \pm 0.01$ & $0.98 \pm 0.01$ \\
BO-2 & $0.08 \pm 0.02$ & $0.16 \pm 0.01$ & $0.34 \pm 0.04$ & $0.07 \pm 0.02$ \\
Ast-2 & $0.51 \pm 0.06$ & $0.65 \pm 0.06$ & $0.87 \pm 0.06$ & $0.60 \pm 0.02$ \\
\hline
Total & $0.46 \pm 0.03$ & $0.56 \pm 0.02$ & $\textbf{0.67} \pm \textbf{0.02}$ & $0.52 \pm 0.01$ \\
\hline
\end{tabular}
\caption{\textbf{CRL: Isolated Params: Variants with isolated parameters across task} improve performance over the baseline. \bigmoe-Hardcoded works the best.}
\label{tab:crl_isolated}
\end{table}

\begin{figure*}[hbt!]
    \centering
    \includegraphics[width=\textwidth]{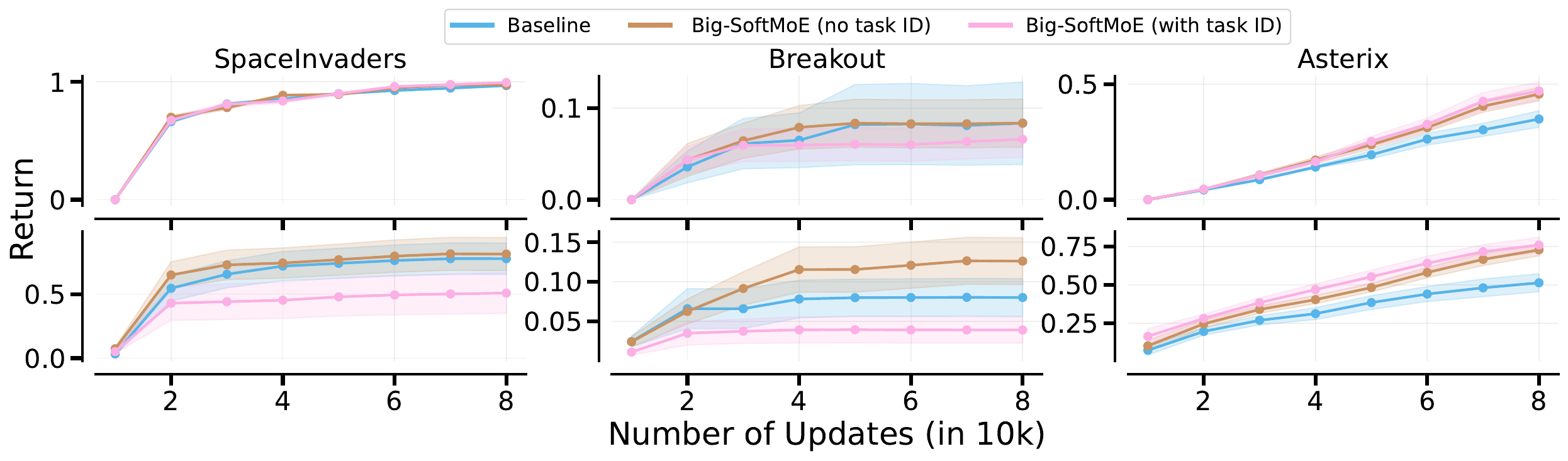}
    \caption{\textbf{CRL: Adding Task ID} as input slightly hurts performance for \bigmoe{}-SoftMoE.}
    \label{fig:crl_task_id}
\end{figure*}

\begin{table}[hbt!]
\centering
\begin{tabular}{|l|c|c|c|}
\hline
Game & Baseline & w/ Task-ID & w/o Task-ID \\
\hline
SI & $0.97 \pm 0.01$ & $0.98 \pm 0.01$ & $0.99 \pm 0.01$ \\
BO & $0.08 \pm 0.05$ & $0.08 \pm 0.03$ & $0.07 \pm 0.02$ \\
Ast & $0.35 \pm 0.04$ & $0.46 \pm 0.03$ & $0.47 \pm 0.04$ \\
SI-2 & $0.78 \pm 0.12$ & $0.81 \pm 0.13$ & $0.51 \pm 0.16$ \\
BO-2 & $0.08 \pm 0.02$ & $0.13 \pm 0.03$ & $0.04 \pm 0.02$ \\
Ast-2 & $0.51 \pm 0.06$ & $0.73 \pm 0.04$ & $0.76 \pm 0.05$ \\
\hline
Total & $0.46 \pm 0.03$ & $\textbf{0.53} \pm \textbf{0.02}$ & $0.47 \pm 0.02$ \\
\hline
\end{tabular}
\caption{\textbf{CRL: Adding Task ID} as input slightly hurts performance for \bigmoe{}-SoftMoE.}
\label{tab:crl_task_id}
\end{table}

\begin{figure*}[hbt!]
    \centering
    \includegraphics[width=\textwidth]{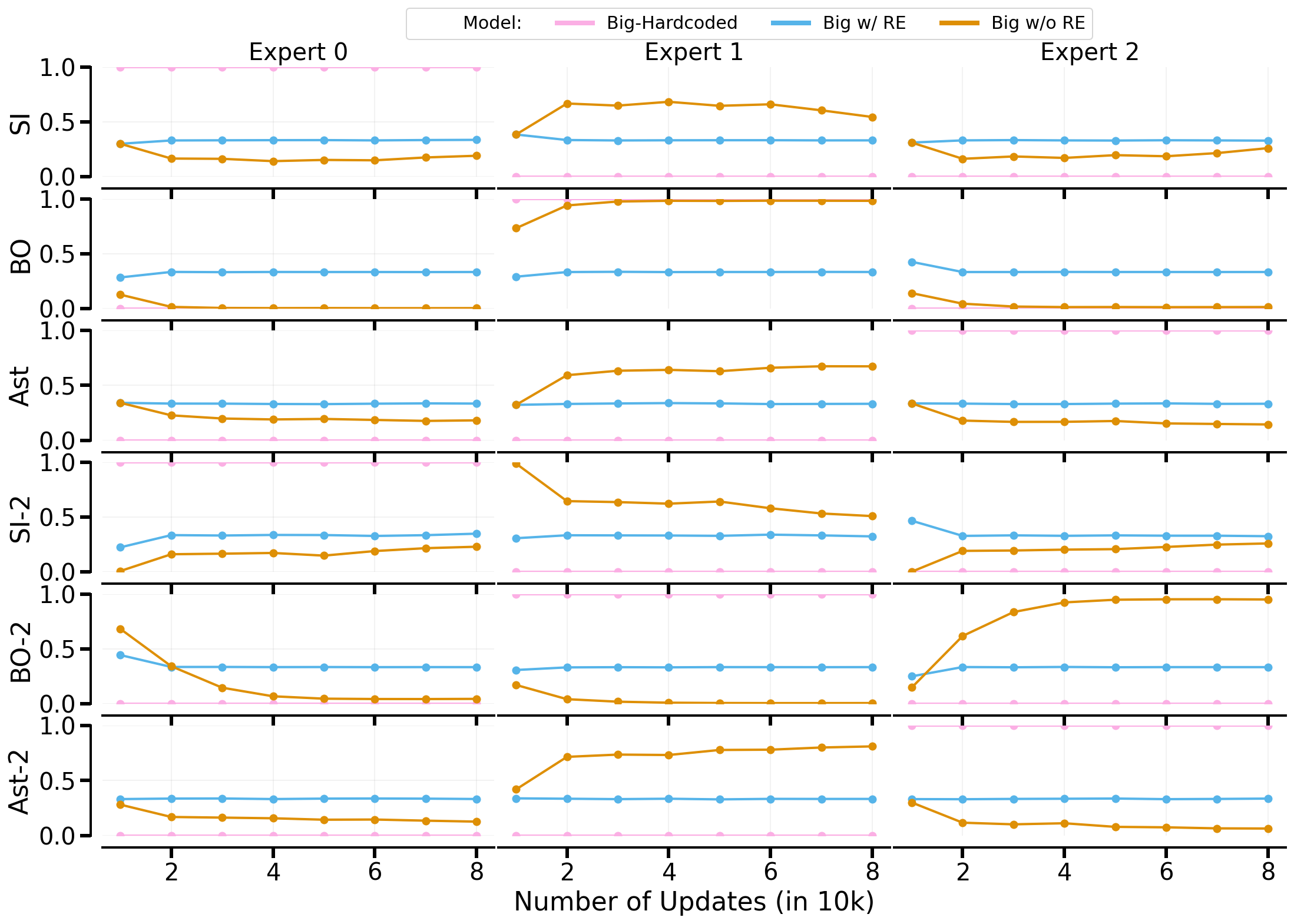}
    \caption{CRL: \bigmoe-SoftMoE also specialises in the CRL setting.}
    \label{fig:crl_big_expert}
\end{figure*}

\begin{table}[hbt!]
\centering
\begin{tabular}{|l|c|c|c|c|}
\hline
Game & Baseline & Big-Hardcoded & Big-SoftGradMoE & Big-SoftMoE \\
\hline
SI & $0.97 \pm 0.01$ & $0.95 \pm 0.00$ & $0.99 \pm 0.01$ & $0.98 \pm 0.01$ \\
BO & $0.08 \pm 0.05$ & $0.32 \pm 0.03$ & $0.12 \pm 0.03$ & $0.08 \pm 0.03$ \\
Ast & $0.35 \pm 0.04$ & $0.56 \pm 0.04$ & $0.45 \pm 0.04$ & $0.46 \pm 0.03$ \\
SI-2 & $0.78 \pm 0.12$ & $0.98 \pm 0.01$ & $0.79 \pm 0.13$ & $0.81 \pm 0.13$ \\
BO-2 & $0.08 \pm 0.02$ & $0.34 \pm 0.04$ & $0.14 \pm 0.03$ & $0.13 \pm 0.03$ \\
Ast-2 & $0.51 \pm 0.06$ & $0.87 \pm 0.06$ & $0.69 \pm 0.06$ & $0.73 \pm 0.04$ \\
\hline
Total & $0.46 \pm 0.03$ & $\textbf{0.67} \pm \textbf{0.02}$ & $0.53 \pm 0.02$ & $0.53 \pm 0.02$ \\
\hline
\end{tabular}
\caption{\textbf{Performance of algorithms across games} with total performance.}
\label{tab:crl_softgradmoe}
\end{table}

\begin{figure*}[hbt!]
    \centering
    \includegraphics[width=\textwidth]{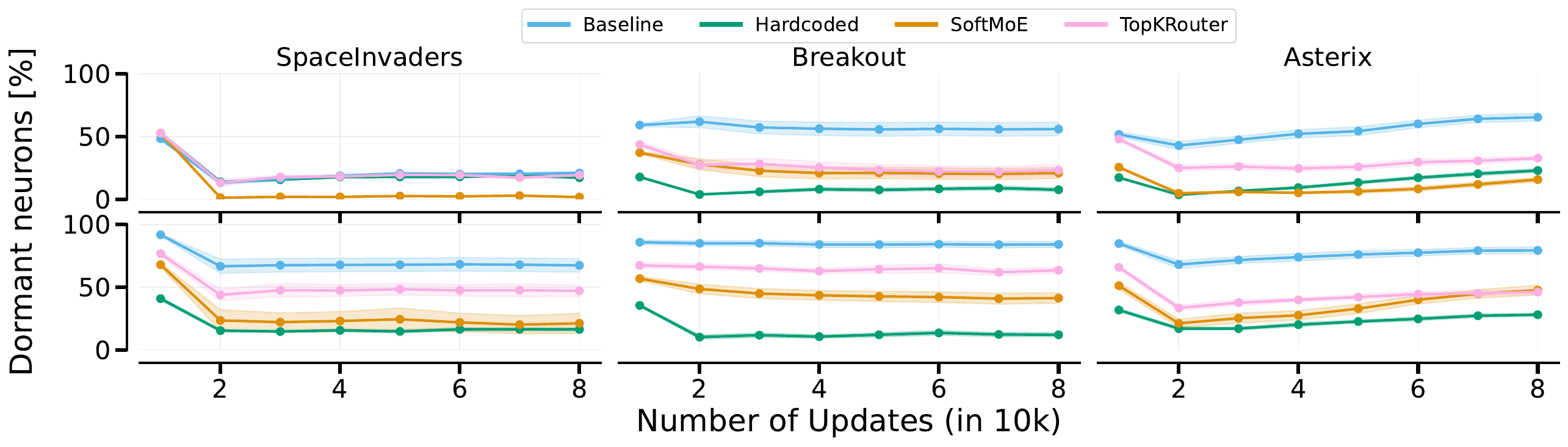}
    %continual_rl/dormant_neurons/crl_dormant.png}
    \caption{\textbf{CRL: MoE variants have less dormant neurons} than the baseline without MoE modules.}
    \label{fig:crl_dormant}
\end{figure*}

\begin{table}[hbt!]
\centering
\begin{tabular}{|l|c|c|c|c|}
\hline
Game & Baseline & \bigmoe{}-Hardcoded & \bigmoe{}-TopK & \bigmoe{}-SoftMoE \\
\hline
SI & $20.86 \pm 1.41$ & $17.19 \pm 0.94$ & $19.22 \pm 1.78$ & $1.72 \pm 0.45$ \\
BO & $55.94 \pm 5.38$ & $7.58 \pm 1.44$ & $23.20 \pm 4.71$ & $20.86 \pm 4.34$ \\
Ast & $65.31 \pm 3.13$ & $22.89 \pm 1.49$ & $32.73 \pm 1.72$ & $15.70 \pm 1.65$ \\
SI-2 & $67.34 \pm 5.38$ & $16.41 \pm 1.22$ & $47.11 \pm 4.49$ & $21.17 \pm 7.97$ \\
BO-2 & $84.06 \pm 2.41$ & $12.11 \pm 1.61$ & $63.52 \pm 2.47$ & $41.33 \pm 4.19$ \\
Ast-2 & $79.30 \pm 2.65$ & $28.05 \pm 1.27$ & $46.17 \pm 1.76$ & $47.58 \pm 3.94$ \\
\hline
Total & $62.14 \pm 1.53$ & $\textbf{17.37} \pm \textbf{0.83}$ & $38.66 \pm 1.59$ & $24.73 \pm 1.58$ \\
\hline
\end{tabular}
\caption{\textbf{CRL Dormant Neurons for \bigmoe{} router variants}. The hardcoded variant has the least dormant neurons}
\label{tab:crl_dormant}
\end{table}

\clearpage
\subsection{Hard-switching based on Gradient Similarity}
\begin{figure*}[hbt!]
    \centering
    \includegraphics[width=\textwidth]{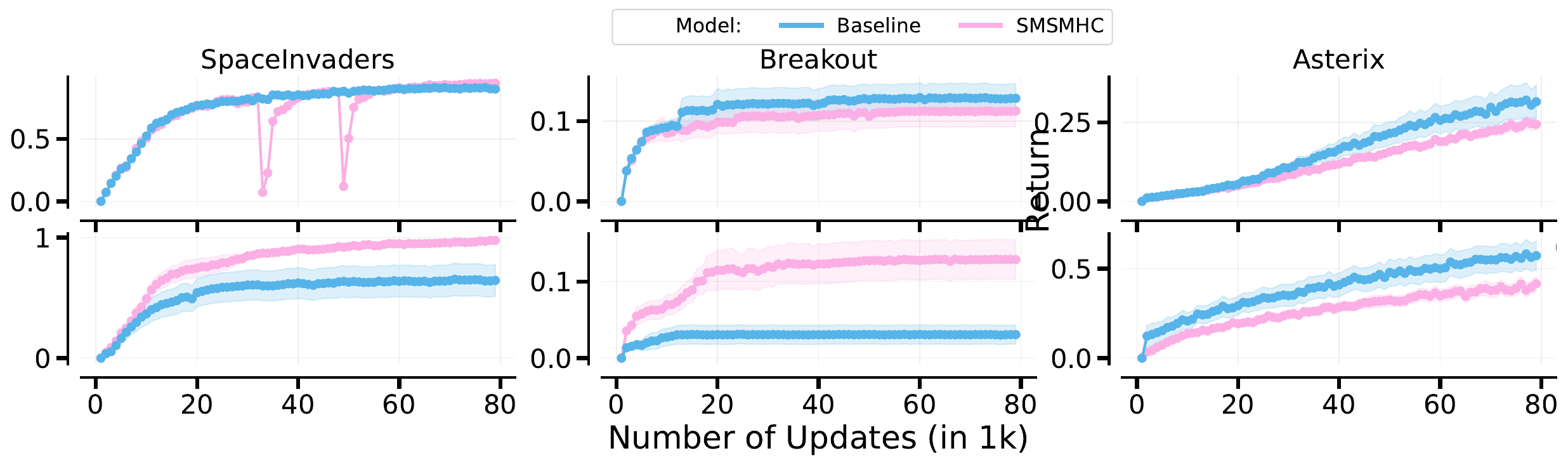}
    \caption{CRL: As shown here with SMSMHC, finding a correct threshold for gradient switching proves difficult, as the experts might switch too early, as in this case it already switches twice during SpaceInvaders (see the dips)}
    \label{fig:crl_grad_switch}
\end{figure*}

We also attempt to route when the gradient similarity drops below a threshold. However, this proved difficult as the thresholds depend on the architecture and expert might switch too early, as shown in \autoref{fig:crl_grad_switch}.

\begin{table}[hbt!]
\centering
\begin{tabular}{|l|c|c|c|c|}
\hline
Game & Baseline & \bigmoe{}-Hardcoded & \bigmoe{}-TopK & \bigmoe{}-SoftMoE \\
\hline
SI & $21.95 \pm 1.15$ & $22.34 \pm 1.12$ & $19.61 \pm 2.10$ & $2.89 \pm 0.40$ \\
BO & $22.50 \pm 3.02$ & $7.89 \pm 1.21$ & $20.39 \pm 1.36$ & $3.91 \pm 1.23$ \\
Ast & $72.66 \pm 1.59$ & $29.14 \pm 1.11$ & $44.30 \pm 2.35$ & $26.02 \pm 1.21$ \\
\hline
Total & $39.04 \pm 0.88$ & $19.79 \pm 0.92$ & $28.10 \pm 1.81$ & $\textbf{10.94} \pm \textbf{0.77}$ \\
\hline
\end{tabular}
\caption{\textbf{MTRL Dormant Neurons for \bigmoe{} router variants}. \bigmoe{}-SoftMoE has the least dormant neurons.}
\label{tab:mtrl_dormant}
\end{table}
\clearpage
\section{Multi-Task RL}
\label{app:mtrl1}
\begin{figure*}[hbt!]
    \centering
    \includegraphics[width=\textwidth]{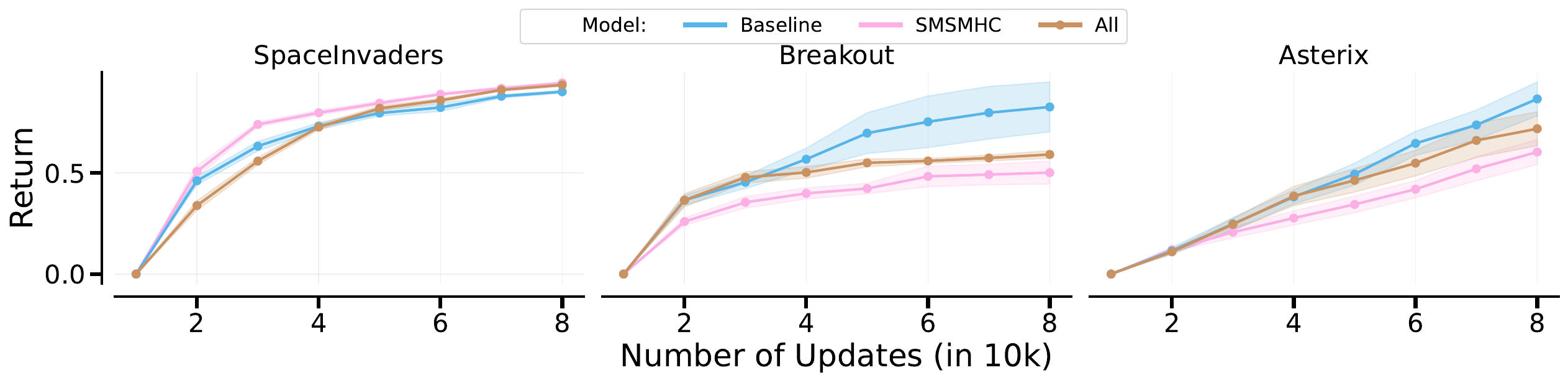}
    %using softMOE
    \caption{\textbf{MTRL: SMSMHC does not improve performance over \allmoe}, suggesting that extreme specialisation in the last layer is not necessarily helpful.}
    \label{fig:mtrl_smsmhc}
\end{figure*}

\paragraph{Combining router learning with some task-specialized layers via Hardcoding.}
% Do we expect final layer specialisation to help? -> We introduce SMSMHC (SoftMoE, SoftMoE, Hardcoded Router), where first two layers are learned SoftMoE layers and last layer is a Hardcoded Router -> Doesn't help performance
We perform preliminary tests how enforced task-specialization in the final layer affects performance. 
For this, we introduce a new architecture termed SMSMHC (SoftMoE, SoftMoE, Hardcoded Router).
%to test whether final layer specialization can be beneficial. 
This architecture consists of two initial layers of learned SoftMoE and a final layer with a Hardcoded Router. Contrary to expectations, SMSMHC does not yield a performance improvement (0.56 $\pm$ 0.02 and 0.53 $\pm$ 0.01), as shown in \autoref{fig:mtrl_smsmhc}, leading to questions about the value of specialization in this context.

\begin{table}[hbt!]
\centering
\begin{tabular}{|l|c|c|c|}
\hline
Game & Baseline & All & SMSMHC \\
\hline
SI & $0.90 \pm 0.01$ & $0.94 \pm 0.01$ & $0.95 \pm 0.00$ \\
BO & $0.46 \pm 0.07$ & $0.33 \pm 0.01$ & $0.28 \pm 0.03$ \\
Ast & $0.51 \pm 0.05$ & $0.43 \pm 0.05$ & $0.36 \pm 0.04$ \\
\hline
Total & $\textbf{0.63} \pm \textbf{0.03}$ & $0.56 \pm 0.02$ & $0.53 \pm 0.01$ \\
\hline
\end{tabular}
%using softMOE
\caption{\textbf{MTRL: SMSMHC does not improve performance over \allmoe}, suggesting that extreme specialisation in the last layer is not necessarily helpful.}
\label{tab:mtrl_smsmhc}
\end{table}

\begin{figure*}[hbt!]
    \centering
    \includegraphics[width=\textwidth]{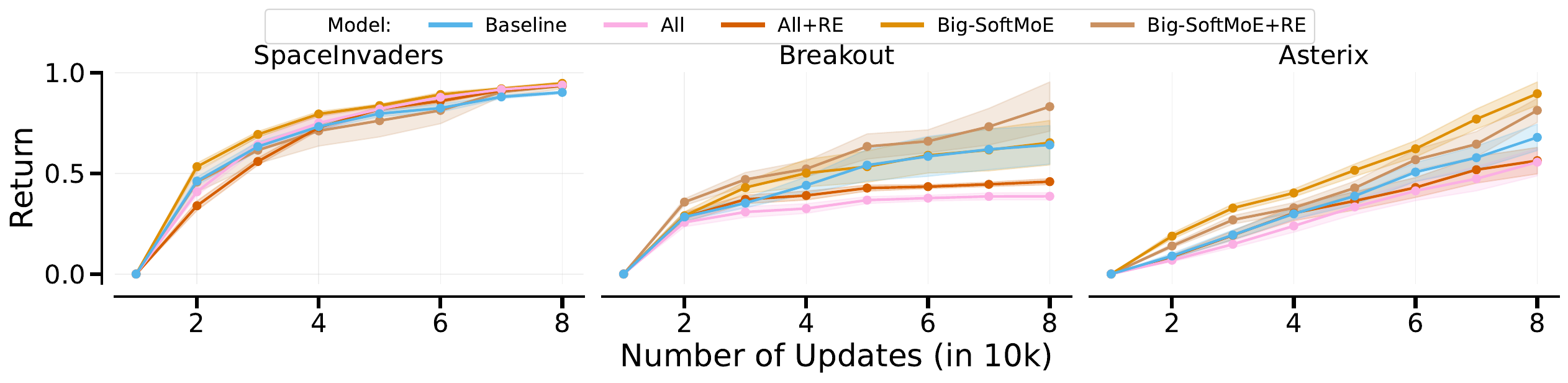}
    %multi_task/regularisation/mtrl_re.png}
    \caption{\textbf{MTRL: Regularising the entropy does not affect performance significantly}, suggesting that specialisation only plays a limited role for performance.}
    \label{fig:mtrl_re}
\end{figure*}

\begin{table}[hbt!]
\centering
\begin{tabular}{|l|c|c|c|c|c|}
\hline
Game & Baseline & All w/o RE & All w/ RE & Big w/o RE & Big w/ RE \\
\hline
SI & $0.90 \pm 0.01$ & $0.94 \pm 0.01$ & $0.94 \pm 0.01$ & $0.95 \pm 0.01$ & $0.93 \pm 0.01$ \\
BO & $0.46 \pm 0.07$ & $0.28 \pm 0.01$ & $0.33 \pm 0.01$ & $0.47 \pm 0.08$ & $0.60 \pm 0.09$ \\
Ast & $0.51 \pm 0.05$ & $0.42 \pm 0.05$ & $0.43 \pm 0.05$ & $0.68 \pm 0.04$ & $0.62 \pm 0.05$ \\
\hline
Total & $0.63 \pm 0.03$ & $0.55 \pm 0.02$ & $0.56 \pm 0.02$ & $\textbf{0.70} \pm \textbf{0.02}$ & $\textbf{0.72} \pm \textbf{0.03}$ \\
\hline
\end{tabular}
\caption{MTRL: Regularising the entropy does not affect performance significantly, suggesting that specialisation only plays a limited role for performance.}
\label{tab:mtrl_re}
\end{table}

\begin{figure*}[hbt!]
    \centering
    \includegraphics[width=\textwidth]{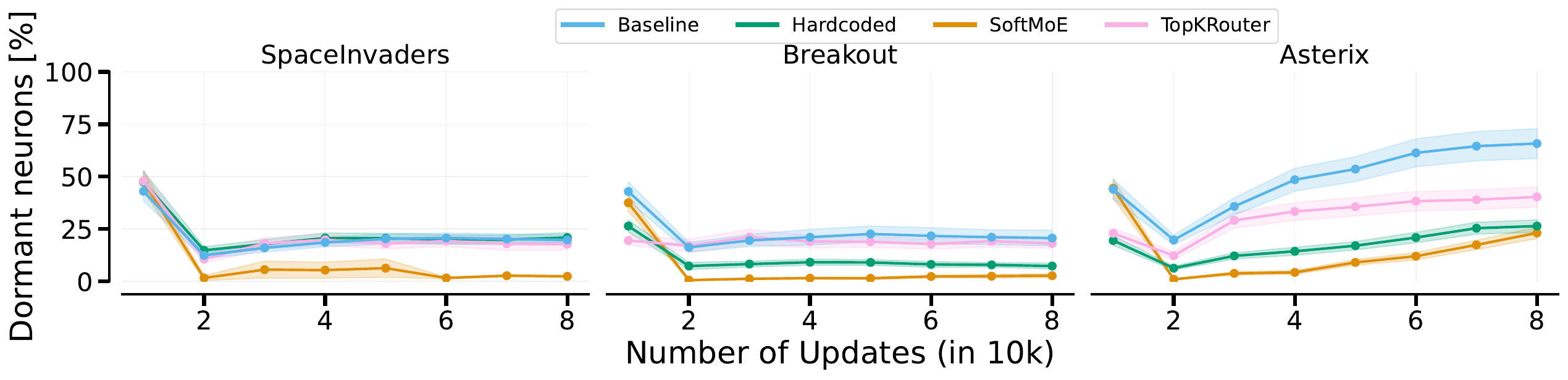}
    \caption{\textbf{MTRL: Generally, dormant neurons are lower} when using \bigmoe variants.}
    \label{fig:mtrl_dormant}
\end{figure*}

\clearpage
\begin{figure*}[hbt!]
    \centering
    \includegraphics[width=\textwidth]{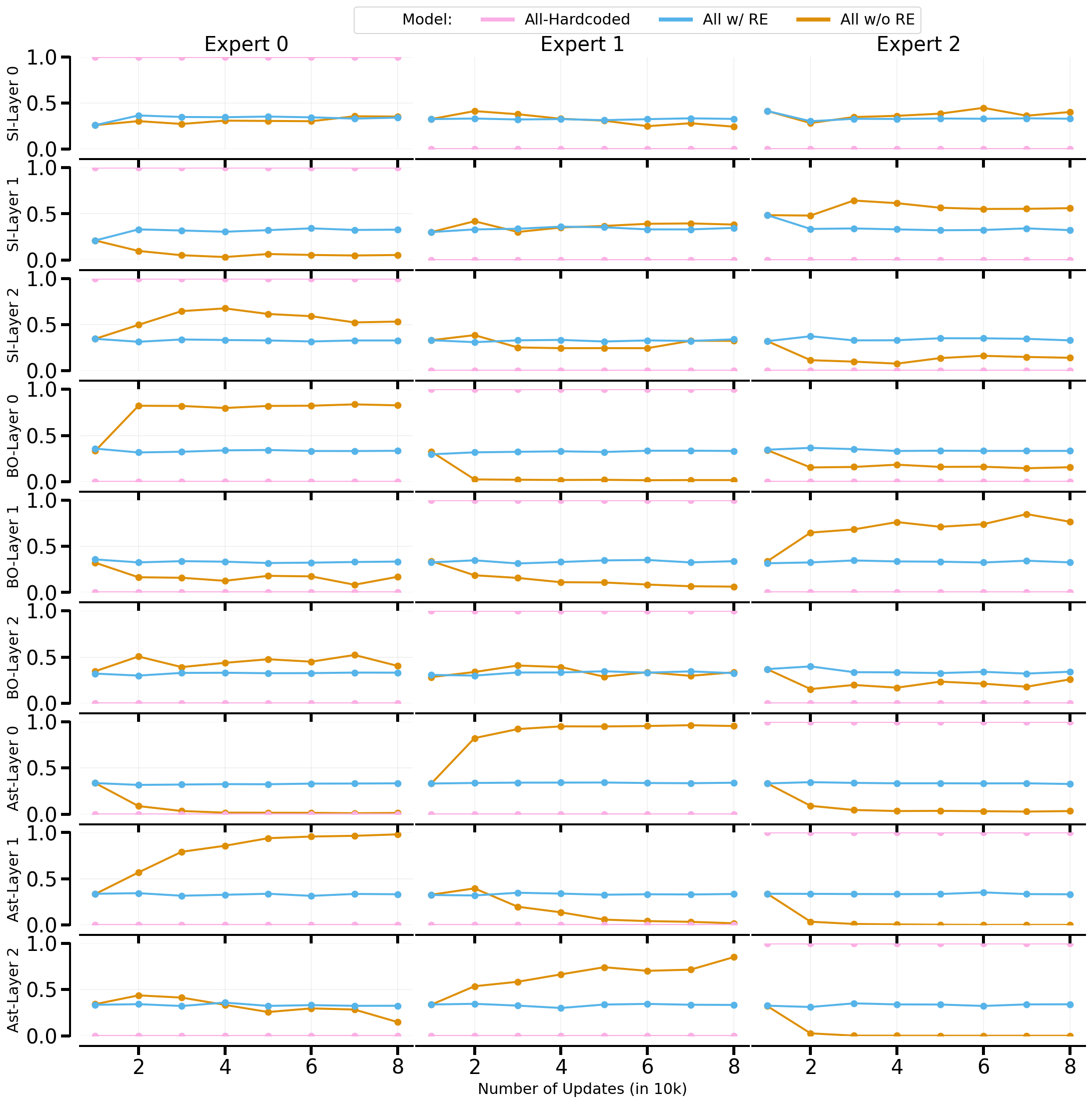}
    \caption{\textbf{MTRL:} Row 1-3 is Layer 1-3 when playing SpaveInvaders, row 4-6, is layer 1-3 when playing Breakout, row 7-9 is layer 1-3 when playing Asterix.}
    \label{fig:mtrl_expert}
\end{figure*}

\clearpage
\section{Actor/Critic Ablation}
\label{app:ac_ablation}
\begin{figure*}[hbt!]
    \centering
    \includegraphics[width=\textwidth]{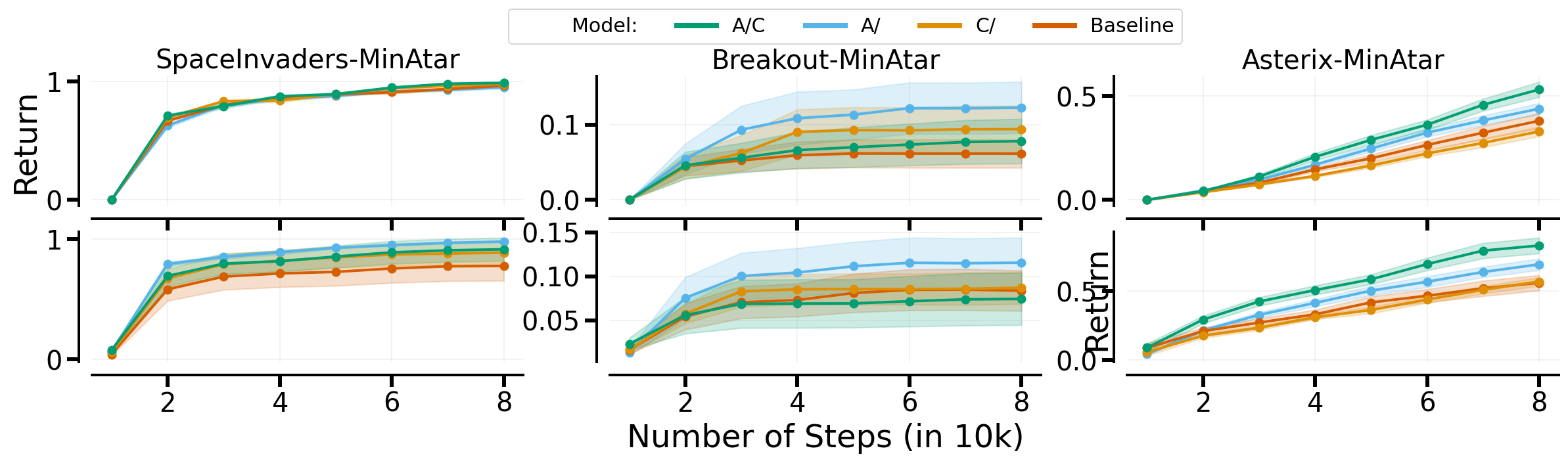}
    \caption{\textbf{CRL: In general, having an MoE module in the actor seems to be most helpful for performance}, whereas the critic MoE module does not improve performance significantly. We used the \bigmoe variant for the ablation.}
    \label{fig:crl_ac2}
\end{figure*}

\begin{table}[hbt!]
\centering
\begin{tabular}{|l|c|c|c|c|}
\hline
Game & Baseline & /C & A/ & A/C \\
\hline
SI & $0.96 \pm 0.00$ & $0.98 \pm 0.01$ & $0.95 \pm 0.01$ & $0.99 \pm 0.01$ \\
BO & $0.06 \pm 0.02$ & $0.09 \pm 0.03$ & $0.12 \pm 0.03$ & $0.08 \pm 0.03$ \\
Ast & $0.38 \pm 0.04$ & $0.33 \pm 0.02$ & $0.44 \pm 0.03$ & $0.53 \pm 0.04$ \\
SI-2 & $0.78 \pm 0.12$ & $0.88 \pm 0.09$ & $0.97 \pm 0.01$ & $0.91 \pm 0.10$ \\
BO-2 & $0.08 \pm 0.02$ & $0.09 \pm 0.02$ & $0.12 \pm 0.03$ & $0.07 \pm 0.03$ \\
Ast-2 & $0.56 \pm 0.06$ & $0.56 \pm 0.03$ & $0.71 \pm 0.05$ & $0.81 \pm 0.05$ \\
\hline
\end{tabular}
\caption{\textbf{CRL: In general, having an MoE module in the actor seems to be most helpful for performance}, whereas the critic MoE module does not improve performance significantly. We used the \bigmoe variant for the ablation.}
\label{tab:crl_ac}
\end{table}

\begin{figure*}[hbt!]
    \centering
    \includegraphics[width=\textwidth]{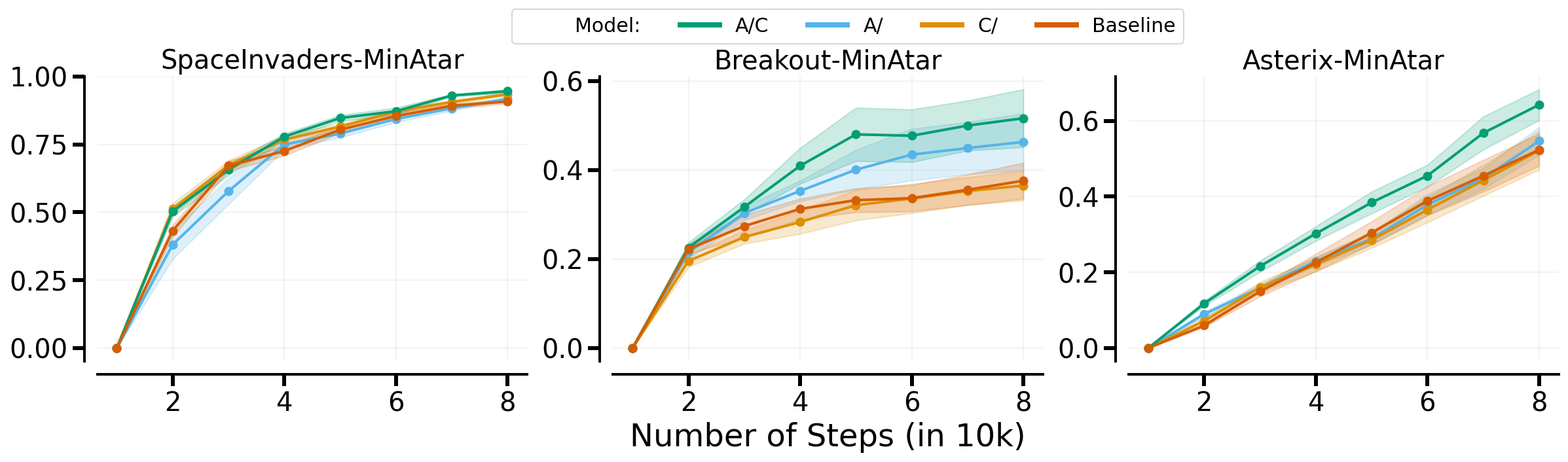}
    \caption{\textbf{MTRL: The combination of Actor and Critic MoE modules appears most beneficial}. We used the \bigmoe variant for the ablation. }
    \label{fig:mtrl_ac}
\end{figure*}

\begin{table}[hbt!]
\centering
\begin{tabular}{|l|c|c|c|c|}
\hline
Game & Baseline & /C & A/ & A/C \\
\hline
SI & $0.91 \pm 0.01$ & $0.93 \pm 0.01$ & $0.92 \pm 0.00$ & $0.95 \pm 0.01$ \\
BO & $0.38 \pm 0.04$ & $0.37 \pm 0.03$ & $0.46 \pm 0.06$ & $0.52 \pm 0.07$ \\
Ast & $0.52 \pm 0.04$ & $0.52 \pm 0.05$ & $0.55 \pm 0.04$ & $0.64 \pm 0.04$ \\
\hline
\textbf{Total} & $0.60 \pm 0.02$ & $0.61 \pm 0.02$ & $0.64 \pm 0.02$ & $\mathbf{0.70 \pm 0.03}$ \\
\hline
\end{tabular}
\caption{\textbf{MTRL: The combination of Actor and Critic MoE modules appears most beneficial}. We used the \bigmoe variant for the ablation.}
\label{tab:mtrl_ac}
\end{table}

\begin{figure*}[hbt!]
    \centering
    \includegraphics[width=\textwidth]{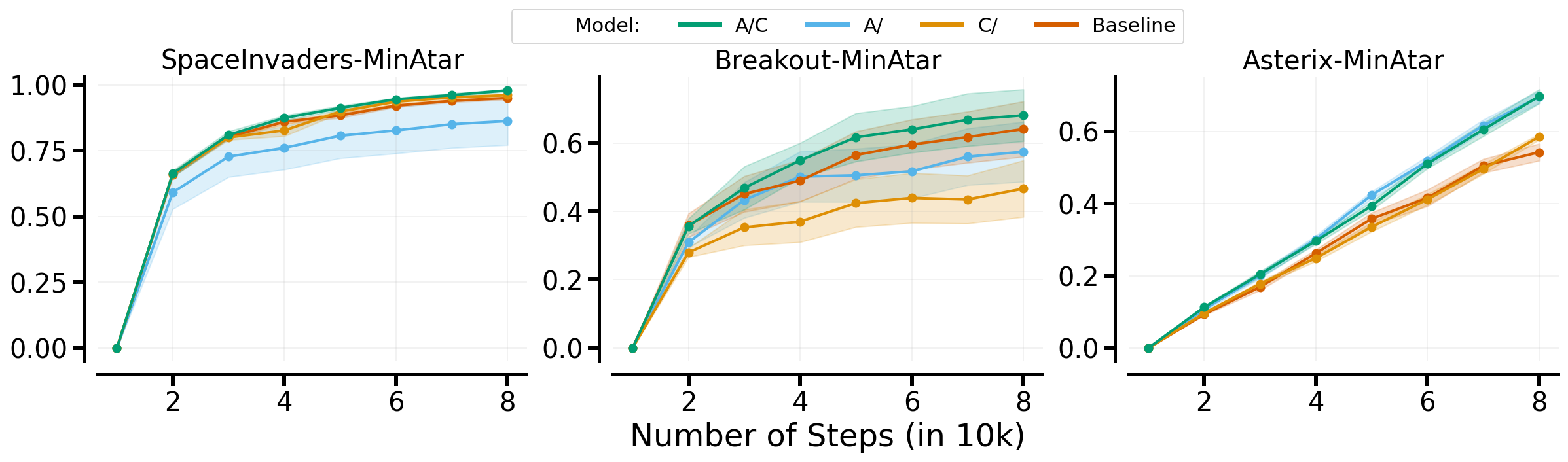}
    \caption{There is no significant difference in using either Actor or Critic, though the combination of both works significantly better than using only the Critic.}
    \label{fig:single_env_ac}
\end{figure*}

\begin{table}[hbt!]
\centering
\begin{tabular}{|l|c|c|c|c|}
\hline
Game & Baseline & A/C & A/ & C/ \\
\hline
SI & $0.95 \pm 0.01$ & $0.98 \pm 0.00$ & $0.86 \pm 0.09$ & $0.96 \pm 0.00$ \\
BO & $0.64 \pm 0.08$ & $0.68 \pm 0.08$ & $0.57 \pm 0.09$ & $0.47 \pm 0.08$ \\
Ast & $0.54 \pm 0.02$ & $0.70 \pm 0.02$ & $0.70 \pm 0.02$ & $0.59 \pm 0.01$ \\
\hline
Total & $\mathbf{0.71 \pm 0.05}$ & $\mathbf{0.79 \pm 0.05}$ & $\mathbf{0.71 \pm 0.07}$ & $0.67 \pm 0.05$ \\
\hline
\end{tabular}
\caption{\textbf{There is no significant difference in using either Actor or Critic}, though the combination of both works significantly better than using only the Critic.}
\label{tab:single_env_ac}
\end{table}

\clearpage
\section{Order Ablation}
\label{app:order_ablation}
\begin{figure*}[hbt!]
    \centering
    \includegraphics[width=\textwidth]{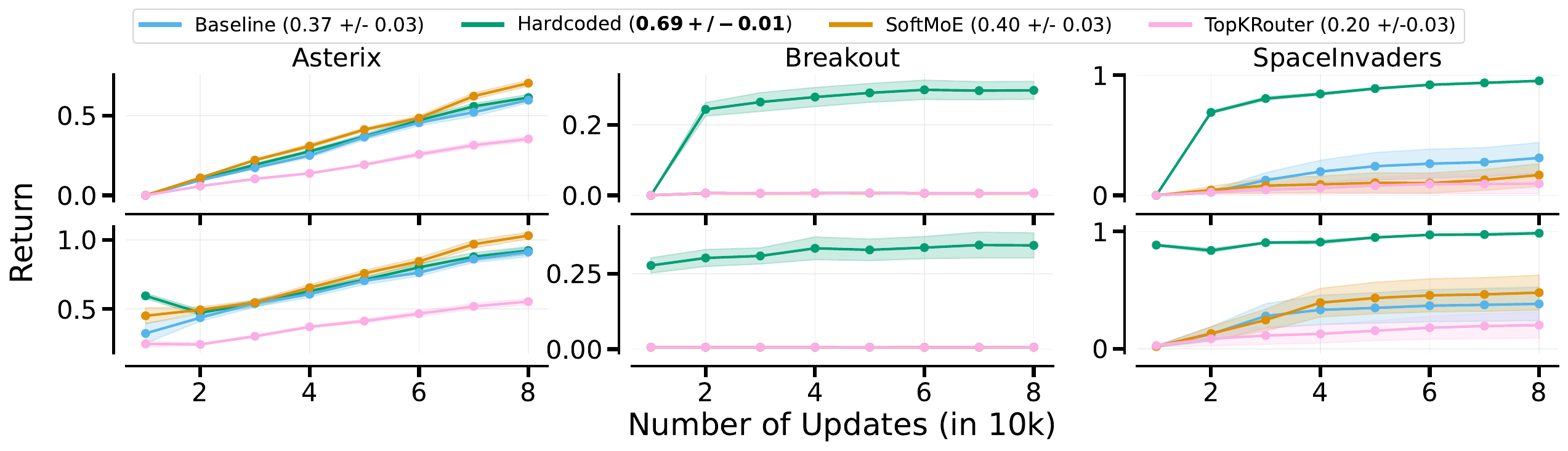}
    \caption{CRL: The order does affect the conclusion for CRL, especially because Breakout performance completely collapses if trained first on Asterix, then Breakout, then SpaceInvaders. Learned routers now do not perform better than the baseline. \bigmoe-Hardcoded still works as expected.}
    \label{fig:crl_order}
\end{figure*}

\begin{table}[hbt!]
\centering
\begin{tabular}{|l|c|c|c|c|}
\hline
Game & Baseline & Big-Hardcoded & Big-TopK & Big-SoftMoE \\
\hline
SI & $0.31 \pm 0.13$ & $0.95 \pm 0.01$ & $0.10 \pm 0.08$ & $0.17 \pm 0.10$ \\
BO & $0.01 \pm 0.00$ & $0.30 \pm 0.03$ & $0.01 \pm 0.00$ & $0.01 \pm 0.00$ \\
Ast & $0.60 \pm 0.02$ & $0.61 \pm 0.02$ & $0.35 \pm 0.02$ & $0.70 \pm 0.02$ \\
SI-2 & $0.38 \pm 0.14$ & $0.98 \pm 0.01$ & $0.20 \pm 0.11$ & $0.48 \pm 0.15$ \\
BO-2 & $0.01 \pm 0.00$ & $0.34 \pm 0.04$ & $0.01 \pm 0.00$ & $0.01 \pm 0.00$ \\
Ast-2 & $0.91 \pm 0.03$ & $0.93 \pm 0.03$ & $0.55 \pm 0.02$ & $1.04 \pm 0.03$ \\
\hline
Total & $0.37 \pm 0.03$ & $\mathbf{0.69 \pm 0.01}$ & $0.20 \pm 0.03$ & $0.40 \pm 0.03$ \\
\hline
\end{tabular}
\caption{CRL: The order does affect the conclusion for CRL, especially because Breakout performance completely collapses if trained first on Asterix, then Breakout, then SpaceInvaders. Learned routers now do not perform better than the baseline. \bigmoe-Hardcoded still works as expected.}
\label{tab:crl_order}
\end{table}

\begin{figure*}[hbt!]
    \centering
    \includegraphics[width=\textwidth]{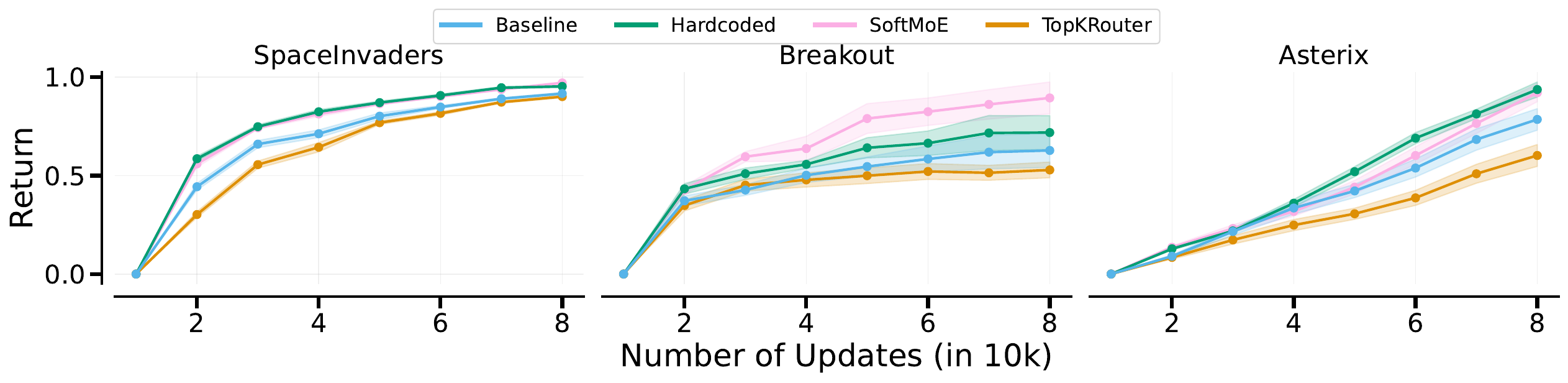}
    %multi_task/order_ablation/mtrl_order.png}
    \caption{\textbf{MTRL: The conclusions do not change when changing the order of training for MTRL}. Learned routers and hardcoded routers perform on par and better than the baseline.}
    \label{fig:mtrl_order}
\end{figure*}

\begin{table}[hbt!]
\centering
\begin{tabular}{|l|c|c|c|c|}
\hline
Game & Baseline & Big-Hardcoded & Big-TopK & Big-SoftMOE \\
\hline
SI & $0.92 \pm 0.01$ & $0.95 \pm 0.00$ & $0.90 \pm 0.01$ & $0.97 \pm 0.00$ \\
BO & $0.34 \pm 0.05$ & $0.39 \pm 0.05$ & $0.29 \pm 0.02$ & $0.49 \pm 0.04$ \\
Ast & $0.51 \pm 0.04$ & $0.61 \pm 0.02$ & $0.39 \pm 0.04$ & $0.60 \pm 0.03$ \\
\hline
Total & $0.59 \pm 0.02$ & \textbf{$0.65 \pm 0.02$} & $0.53 \pm 0.01$ & \textbf{$0.68 \pm 0.02$} \\
\hline
\end{tabular}
\caption{\textbf{MTRL: The conclusions do not change when changing the order of training for MTRL}. Learned routers and hardcoded routers perform on par and better than the baseline.}
\label{tab:mtrl_order}
\end{table}

\clearpage
\section{MTRL - More Results}
\label{app:mtrl_more_results}
\begin{figure*}[hbt!]
    \centering
    \includegraphics[width=\textwidth]{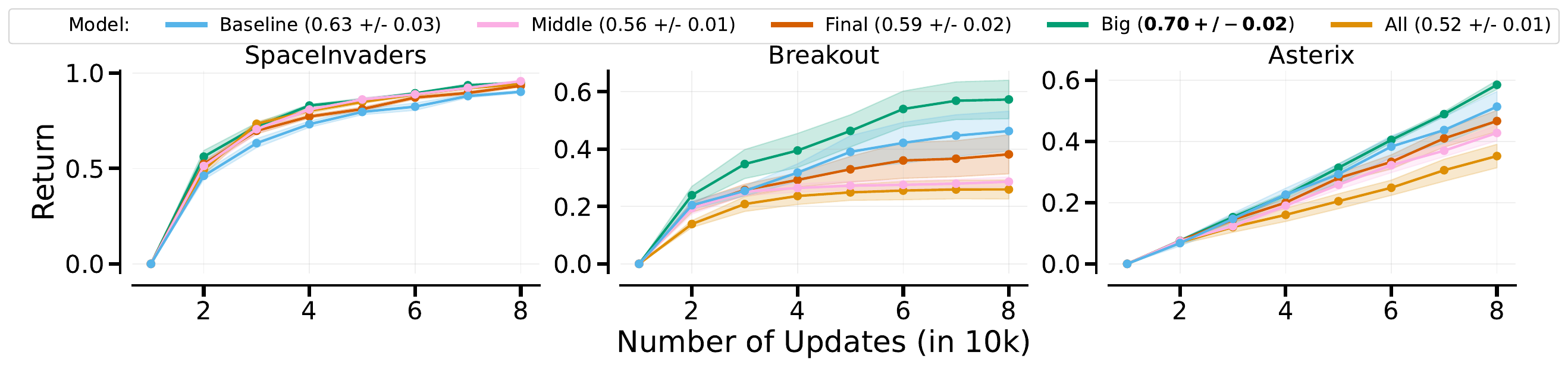}
    %multi_task/hardcodedrouter/mtrl_hardcoded.png}
    \caption{\textbf{MTRL: All Hardcoded architectures, \bigmoe-Hardcoded works best}. It is unclear why \allmoe does not perform as well as \bigmoe, though we hypothesise it is due to suboptimal hyperparameters}
    \label{fig:mtrl_hardcoded}
\end{figure*}

\begin{table}[hbt!]
\centering
\begin{tabular}{|l|c|c|c|c|c|}
\hline
Game & Baseline & \allmoe & \bigmoe & \finalmoe & \middlemoe \\
\hline
SI & $0.90 \pm 0.01$ & $0.94 \pm 0.01$ & $0.95 \pm 0.01$ & $0.93 \pm 0.01$ & $0.96 \pm 0.01$ \\
BO & $0.46 \pm 0.07$ & $0.26 \pm 0.03$ & $0.57 \pm 0.07$ & $0.38 \pm 0.07$ & $0.29 \pm 0.02$ \\
Ast & $0.51 \pm 0.05$ & $0.35 \pm 0.04$ & $0.59 \pm 0.01$ & $0.47 \pm 0.03$ & $0.43 \pm 0.03$ \\
\hline
Total & $0.63 \pm 0.03$ & $0.52 \pm 0.01$ & \textbf{$0.70 \pm 0.02$} & $0.59 \pm 0.02$ & $0.56 \pm 0.01$ \\
\hline
\end{tabular}
\caption{\textbf{MTRL: All Hardcoded architectures: \bigmoe-Hardcoded works best}. It is unclear why \allmoe does not perform as well as \bigmoe, though we hypothesise it is due to suboptimal hyperparameters.}
\label{tab:mtrl_hardcoded}
\end{table}

\begin{figure*}[hbt!]
    \centering
    \includegraphics[width=\textwidth]{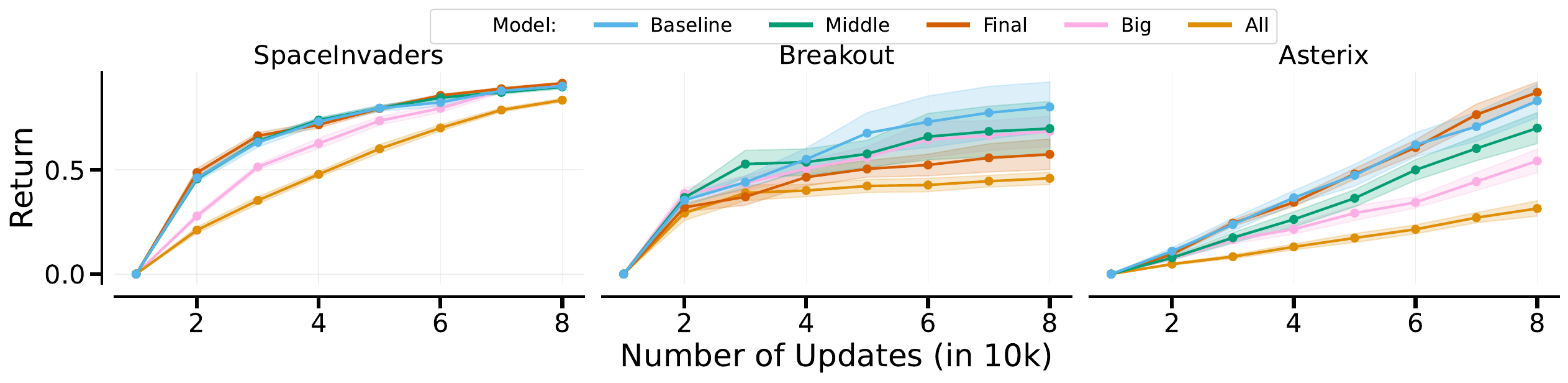}
    %multi_task/topkrouter/mtrl_topk.png}
    \caption{\textbf{MTRL: All TopKRouter architectures:} Generally, TopKRouters perform worse than the baseline.}
    \label{fig:mtrl_topk}
\end{figure*}

\begin{table}[hbt!]
\centering
\begin{tabular}{|l|c|c|c|c|c|}
\hline
Game & Baseline & \allmoe & \bigmoe & \finalmoe & \middlemoe \\
\hline
SI & $0.90 \pm 0.01$ & $0.83 \pm 0.01$ & $0.91 \pm 0.01$ & $0.92 \pm 0.00$ & $0.90 \pm 0.01$ \\
BO & $0.46 \pm 0.07$ & $0.27 \pm 0.02$ & $0.40 \pm 0.04$ & $0.33 \pm 0.04$ & $0.40 \pm 0.07$ \\
Ast & $0.51 \pm 0.05$ & $0.19 \pm 0.02$ & $0.34 \pm 0.03$ & $0.54 \pm 0.03$ & $0.43 \pm 0.05$ \\
\hline
Total & $\mathbf{0.63 \pm 0.03}$ & $0.43 \pm 0.01$ & $0.55 \pm 0.02$ & $\mathbf{0.60 \pm 0.01}$ & $\mathbf{0.58 \pm 0.03}$ \\
\hline
\end{tabular}
\caption{\textbf{MTRL: All TopKRouter architectures:} Generally, TopKRouters perform worse than the baseline.}
\label{tab:mtrl_topk}
\end{table}

\begin{figure*}[hbt!]
    \centering
    \includegraphics[width=\textwidth]{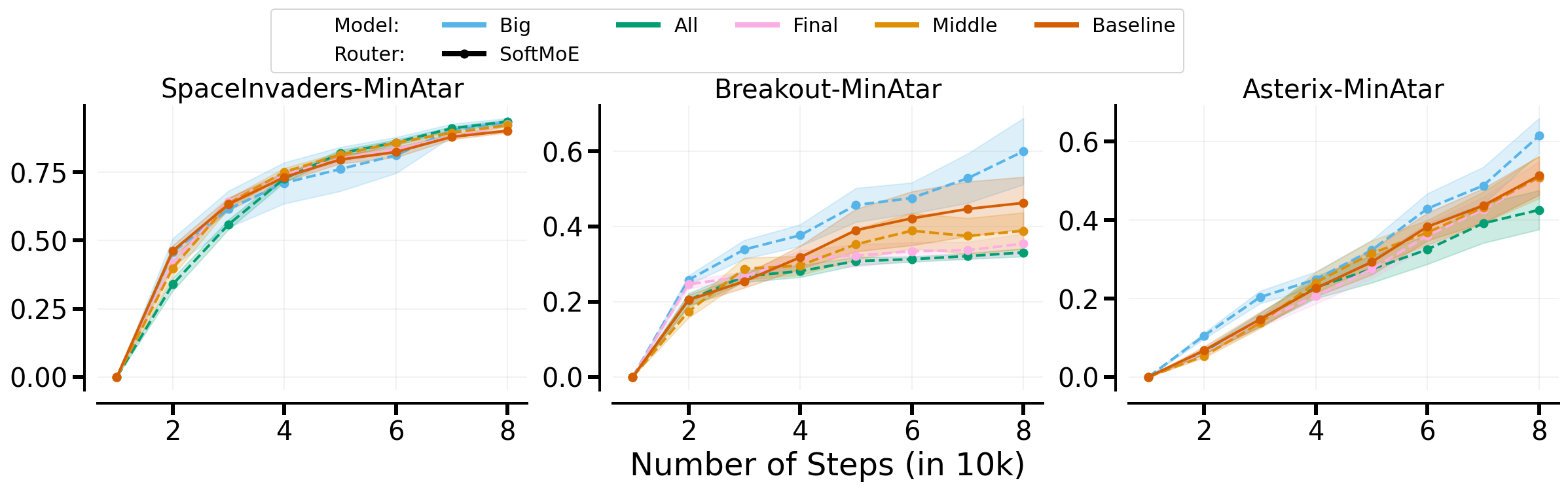}
    \caption{\textbf{MTRL: All SoftMoE Architectures:} Only \bigmoe-SoftMoE performs better than the baseline.}
    \label{fig:mtrl_softmoe}
\end{figure*}

\begin{table}[hbt!]
\centering
\begin{tabular}{|l|c|c|c|c|c|}
\hline
Game & Baseline & \allmoe & \bigmoe & \finalmoe & \middlemoe \\
\hline
SI & $0.90 \pm 0.01$ & $0.94 \pm 0.01$ & $0.93 \pm 0.01$ & $0.92 \pm 0.01$ & $0.92 \pm 0.01$ \\
BO & $0.46 \pm 0.07$ & $0.33 \pm 0.01$ & $0.60 \pm 0.09$ & $0.35 \pm 0.03$ & $0.39 \pm 0.05$ \\
Ast & $0.51 \pm 0.05$ & $0.43 \pm 0.05$ & $0.62 \pm 0.05$ & $0.51 \pm 0.04$ & $0.51 \pm 0.05$ \\
\hline
Total & $0.63 \pm 0.03$ & $0.56 \pm 0.02$ & $\mathbf{0.72 \pm 0.03}$ & $0.60 \pm 0.02$ & $0.61 \pm 0.02$ \\
\hline
\end{tabular}
\caption{\textbf{MTRL: All SoftMoE Architectures:} Only \bigmoe-SoftMoE performs better than the baseline.}
\label{tab:mtrl_softmoe}
\end{table}

\begin{figure*}[hbt!]
    \centering
    \includegraphics[width=\textwidth]{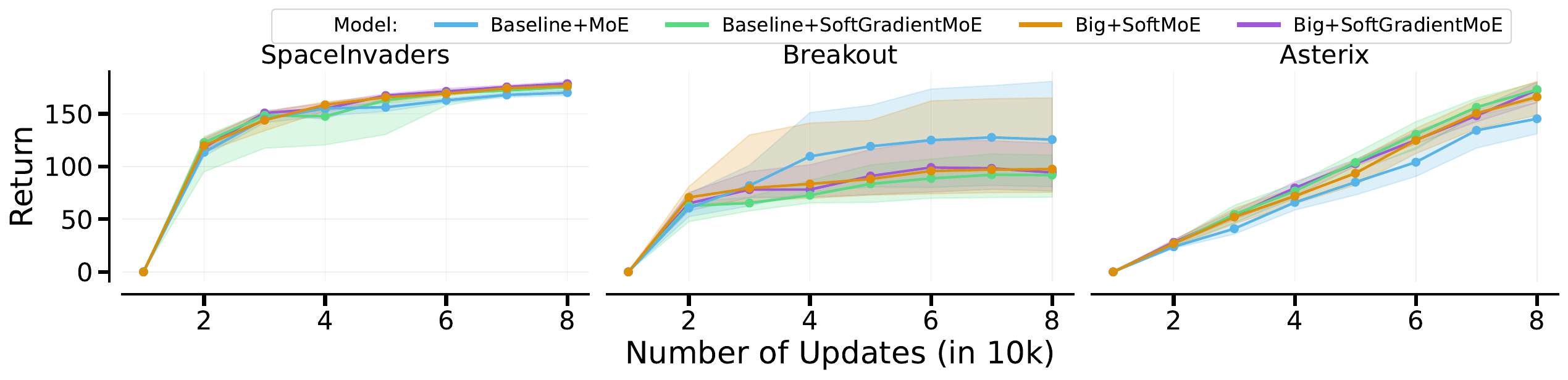}
    %multi_task/softgradientmoe/mtrl_softgrad.png}
        \caption{\textbf{MTRL: Big-SoftGradientMoE vs. Big-SoftMoE}, adding gradient information does not improve performance.}
    \label{fig:mtrl_softgrad}
\end{figure*}

\begin{table}[hbt!]
\centering
\begin{tabular}{|l|c|c|c|c|}
\hline
Game & Baseline & \bigmoe-Hardcoded & \bigmoe-SoftGradientMoE & \bigmoe-SoftMoE \\
\hline
SI & $0.90 \pm 0.01$ & $0.95 \pm 0.01$ & $0.95 \pm 0.00$ & $0.93 \pm 0.01$ \\
BO & $0.46 \pm 0.07$ & $0.57 \pm 0.07$ & $0.51 \pm 0.07$ & $0.60 \pm 0.09$ \\
Ast & $0.51 \pm 0.05$ & $0.59 \pm 0.01$ & $0.66 \pm 0.03$ & $0.62 \pm 0.05$ \\
\hline
Total & $0.63 \pm 0.03$ & $\mathbf{0.70 \pm 0.02}$ & $\mathbf{0.70 \pm 0.03}$ & $\mathbf{0.72 \pm 0.03}$ \\
\hline
\end{tabular}
\caption{\textbf{MTRL: Big-SoftGradientMoE vs. Big-SoftMoE}, adding gradient information does not improve performance.}
\label{tab:mtrl_softgrad}
\end{table}

\clearpage
\section{CRL - More Results}
\label{app:crl_more_results}
\begin{figure*}[hbt!]
    \centering
    \includegraphics[width=\textwidth]{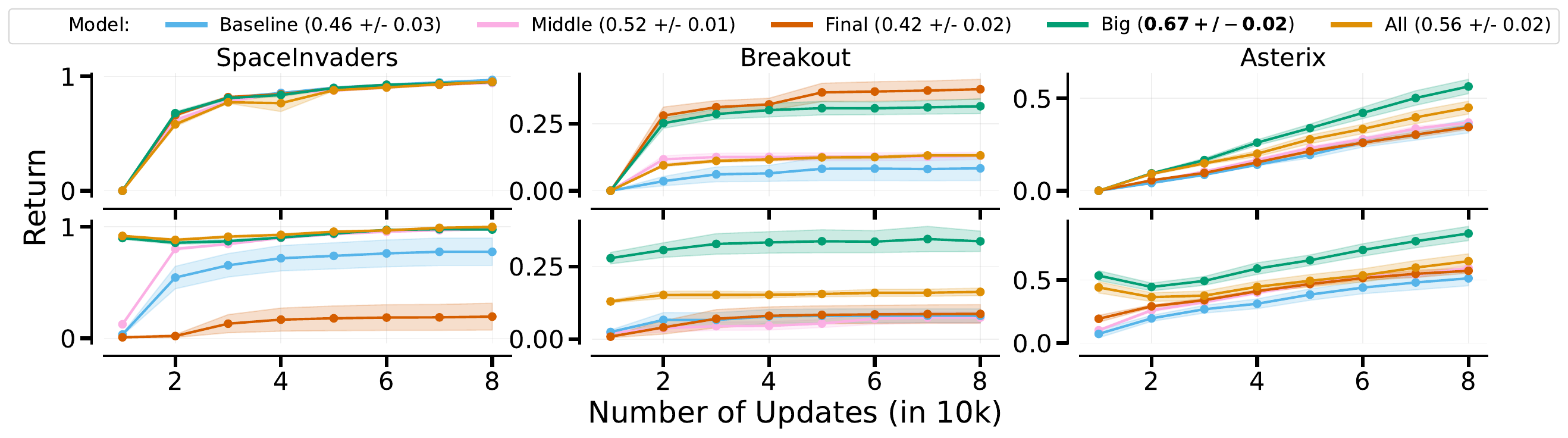}
    %continual_rl/hardcodedrouter/crl_hardcoded.png}
    \caption{\textbf{CRL: All HardcodedRouter Architectures}. \bigmoe-Hardcoded works best. It is unclear why AllLayers performs significantly worse on Breakout, though we hypothesise it is due to suboptimal hyperparameters}
    \label{fig:crl_hardcoded}
\end{figure*}

\begin{table}[hbt!]
\centering
\begin{tabular}{|l|c|c|c|c|c|}
\hline
Game & Baseline & \allmoe & \bigmoe & \finalmoe & \middlemoe \\
\hline
SI   & $0.97 \pm 0.01$ & $0.95 \pm 0.01$ & $0.95 \pm 0.00$ & $0.95 \pm 0.01$ & $0.94 \pm 0.01$ \\
BO   & $0.08 \pm 0.05$ & $0.13 \pm 0.00$ & $0.32 \pm 0.03$ & $0.38 \pm 0.04$ & $0.13 \pm 0.01$ \\
Ast  & $0.35 \pm 0.04$ & $0.45 \pm 0.03$ & $0.56 \pm 0.04$ & $0.34 \pm 0.01$ & $0.37 \pm 0.01$ \\
SI-2 & $0.78 \pm 0.12$ & $1.00 \pm 0.00$ & $0.98 \pm 0.01$ & $0.19 \pm 0.12$ & $0.98 \pm 0.01$ \\
BO-2 & $0.08 \pm 0.02$ & $0.16 \pm 0.01$ & $0.34 \pm 0.04$ & $0.09 \pm 0.03$ & $0.07 \pm 0.02$ \\
Ast-2 & $0.51 \pm 0.06$ & $0.65 \pm 0.06$ & $0.87 \pm 0.06$ & $0.57 \pm 0.02$ & $0.60 \pm 0.02$ \\
\hline
Total & $0.46 \pm 0.03$ & $0.56 \pm 0.02$ & $\mathbf{0.67 \pm 0.02}$ & $0.42 \pm 0.02$ & $0.52 \pm 0.01$ \\
\hline
\end{tabular}
\caption{\textbf{CRL: All HardcodedRouter Architectures}. \bigmoe-Hardcoded works best. It is unclear why AllLayers performs significantly worse on Breakout, though we hypothesise it is due to suboptimal hyperparameters}
\label{tab:crl_hardcoded}
\end{table}

\begin{figure*}[hbt!]
    \centering
    \includegraphics[width=\textwidth]{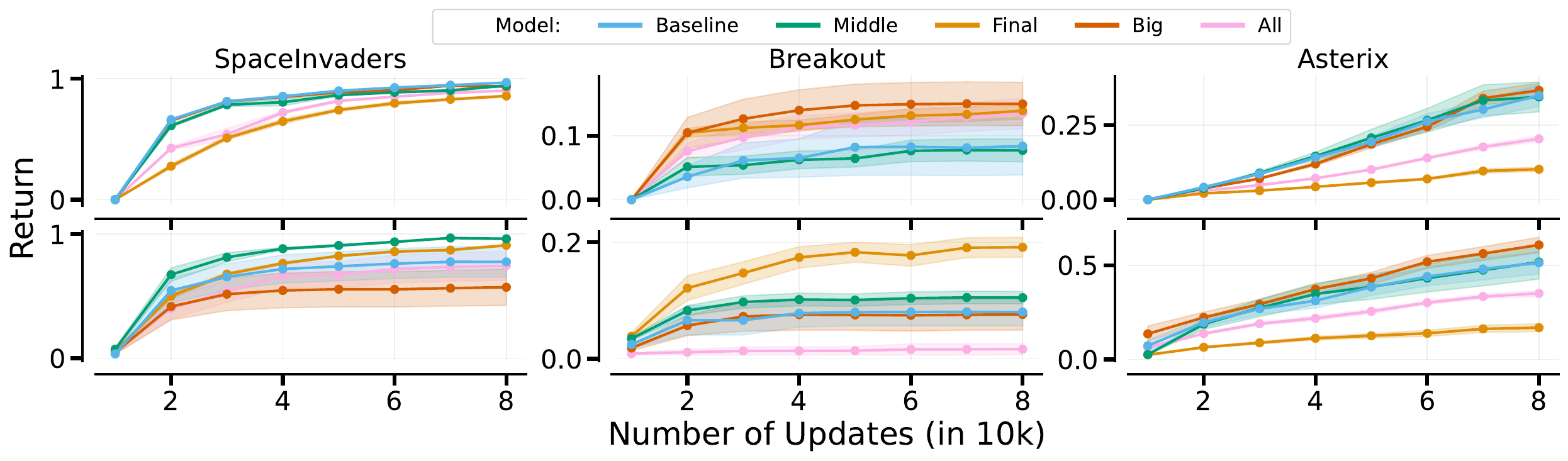}
    %figures/continual_rl/topkrouter/crl_topk.png}
    \caption{\textbf{CRL: All TopK variants} perform worse than the baseline except for \middlemoe.}
    \label{fig:crl_topk}
\end{figure*}

\begin{table}[hbt!]
\centering
\begin{tabular}{|l|c|c|c|c|c|}
\hline
Game & Baseline & \allmoe & \bigmoe & \finalmoe & \middlemoe \\
\hline
SI   & $0.97 \pm 0.01$ & $0.86 \pm 0.01$ & $0.90 \pm 0.01$ & $0.94 \pm 0.01$ & $0.94 \pm 0.01$ \\
BO   & $0.08 \pm 0.05$ & $0.14 \pm 0.01$ & $0.13 \pm 0.02$ & $0.15 \pm 0.03$ & $0.08 \pm 0.02$ \\
Ast  & $0.35 \pm 0.04$ & $0.10 \pm 0.01$ & $0.20 \pm 0.01$ & $0.37 \pm 0.02$ & $0.34 \pm 0.05$ \\
SI-2 & $0.78 \pm 0.12$ & $0.91 \pm 0.01$ & $0.74 \pm 0.12$ & $0.57 \pm 0.15$ & $0.96 \pm 0.01$ \\
BO-2 & $0.08 \pm 0.02$ & $0.19 \pm 0.02$ & $0.02 \pm 0.01$ & $0.08 \pm 0.03$ & $0.10 \pm 0.01$ \\
Ast-2 & $0.51 \pm 0.06$ & $0.17 \pm 0.02$ & $0.35 \pm 0.02$ & $0.61 \pm 0.04$ & $0.52 \pm 0.09$ \\
\hline
Total & $\mathbf{0.46 \pm 0.03}$ & $0.39 \pm 0.01$ & $0.39 \pm 0.02$ & $\mathbf{0.45 \pm 0.02}$ & $\mathbf{0.49 \pm 0.02}$ \\
\hline
\end{tabular}
\caption{\textbf{CRL: All TopK variants} perform worse than the baseline except for \middlemoe.}
\label{tab:crl_topk}
\end{table}

\begin{figure*}[hbt!]
    \centering
    \includegraphics[width=\textwidth]{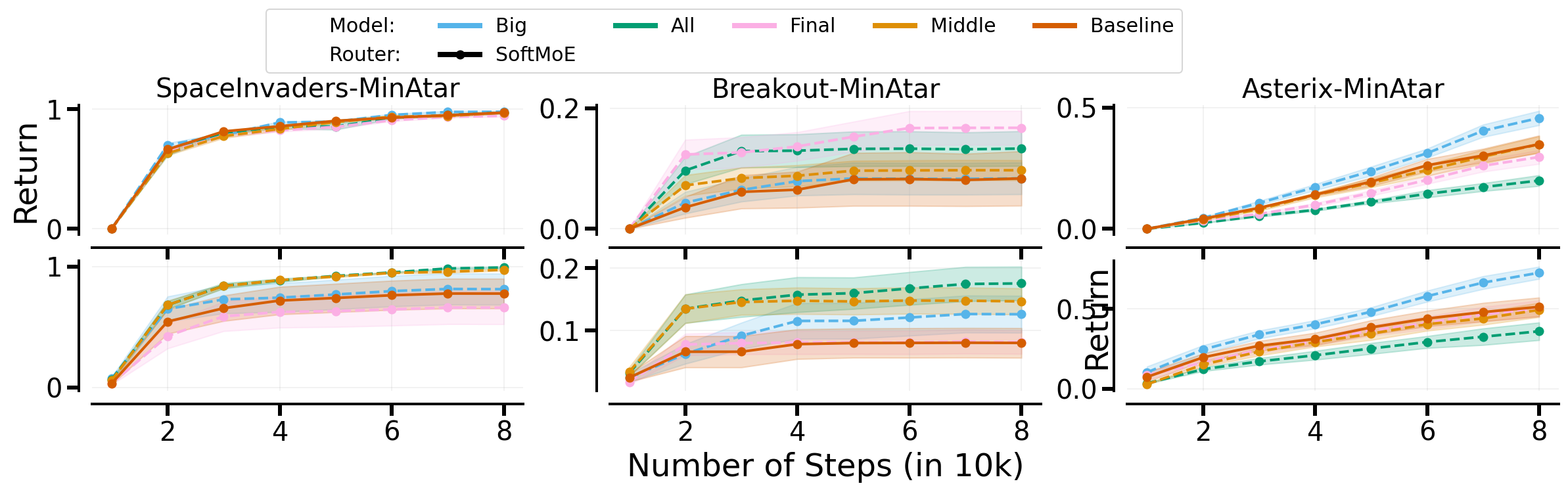}
    \caption{\textbf{CRL: All SoftMoE Architectures}: \bigmoe-SoftMoE is the only variant that performs better than the baseline.}
    \label{fig:crl_softmoe}
\end{figure*}

\begin{table}[hbt!]
\centering
\begin{tabular}{|l|c|c|c|c|c|}
\hline
Game & Baseline & \allmoe & \bigmoe & \finalmoe & \middlemoe \\
\hline
SI   & $0.97 \pm 0.01$ & $0.97 \pm 0.01$ & $0.98 \pm 0.01$ & $0.94 \pm 0.01$ & $0.97 \pm 0.01$ \\
BO   & $0.08 \pm 0.05$ & $0.13 \pm 0.03$ & $0.08 \pm 0.03$ & $0.17 \pm 0.03$ & $0.10 \pm 0.02$ \\
Ast  & $0.35 \pm 0.04$ & $0.20 \pm 0.02$ & $0.46 \pm 0.03$ & $0.30 \pm 0.03$ & $0.35 \pm 0.03$ \\
SI-2 & $0.78 \pm 0.12$ & $0.99 \pm 0.01$ & $0.81 \pm 0.13$ & $0.66 \pm 0.14$ & $0.97 \pm 0.01$ \\
BO-2 & $0.08 \pm 0.02$ & $0.18 \pm 0.03$ & $0.13 \pm 0.03$ & $0.08 \pm 0.02$ & $0.15 \pm 0.02$ \\
Ast-2 & $0.51 \pm 0.06$ & $0.36 \pm 0.06$ & $0.73 \pm 0.04$ & $0.50 \pm 0.05$ & $0.49 \pm 0.05$ \\
\hline
Total & $\mathbf{0.46 \pm 0.03}$ & $0.47 \pm 0.01$ & $\mathbf{0.53 \pm 0.02}$ & $0.44 \pm 0.03$ & $\mathbf{0.50 \pm 0.01}$ \\
\hline
\end{tabular}
\caption{\textbf{CRL: All SoftMoE Architectures:} \bigmoe-SoftMoE is the only variant that performs better than the baseline.}
\label{tab:crl_softmoe}
\end{table}

\begin{figure*}[hbt!]
    \centering
    \includegraphics[width=\textwidth]{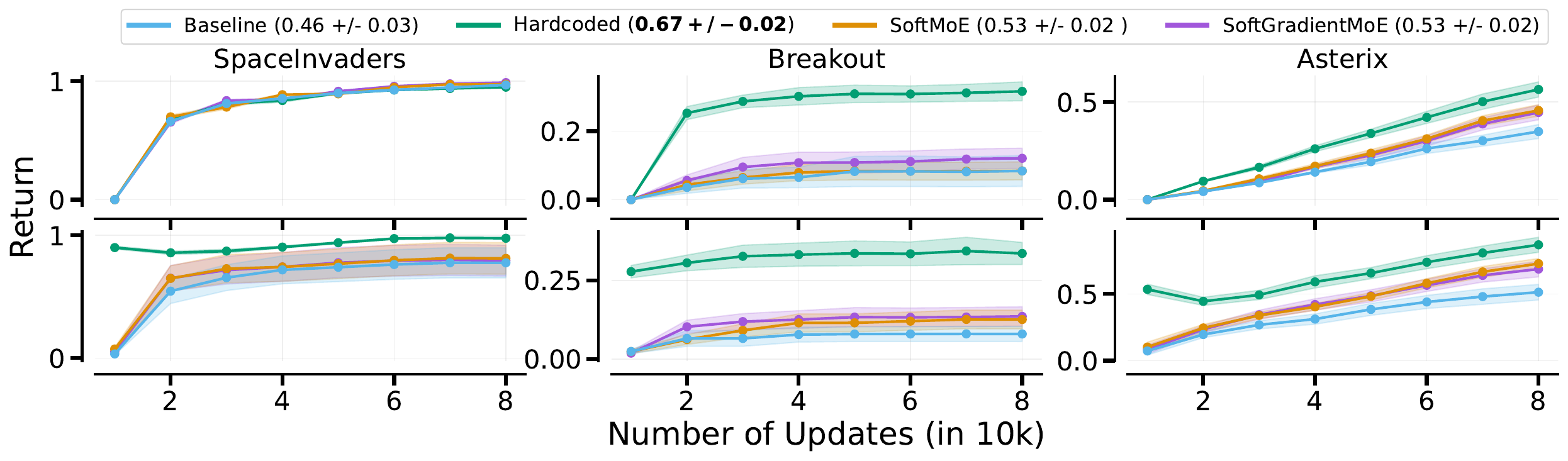}
    %continual_rl/softgradientmoe/crl_softgrad.png}
    \caption{\textbf{CRL: Adding gradient information to the input} does not improve performance significantly.}
    \label{fig:crl_softgrad}
\end{figure*}

\clearpage
\section{Single Env}
\label{app:single_env_more_results}
% \begin{figure*}[hbt!]
%     \centering
%     \includegraphics[width=\textwidth]{figures/single_env/hardcodedrouter/single_env_hardcoded.png}
%     \caption{Single Env: All Hardcoded Architectures}
%     \label{fig:single_env_hardcoded}
% \end{figure*}
\begin{figure*}[hbt!]
    \centering
    \includegraphics[width=0.28\textwidth]{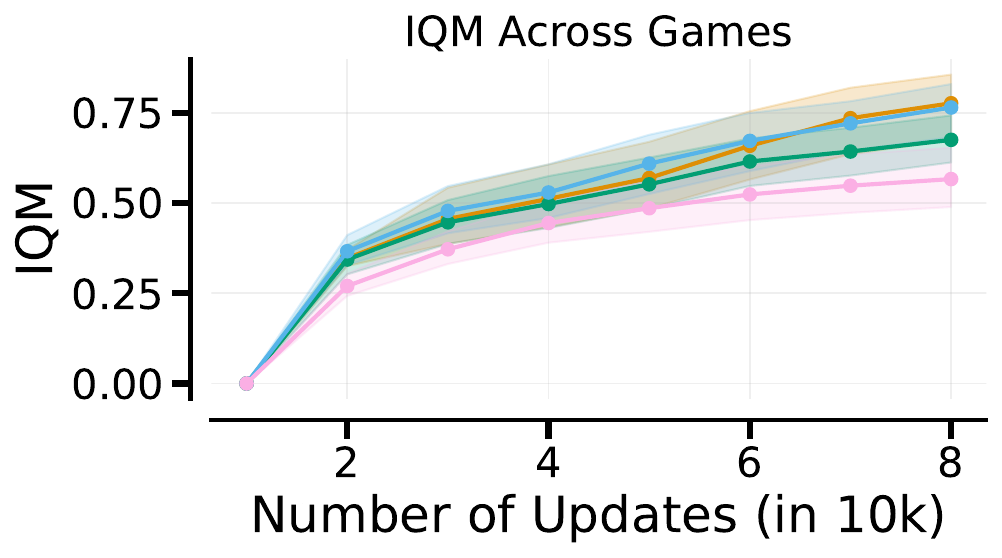}%
       \vline height 75pt depth 0 pt width 0.8 pt
    \includegraphics[width=0.72\textwidth]{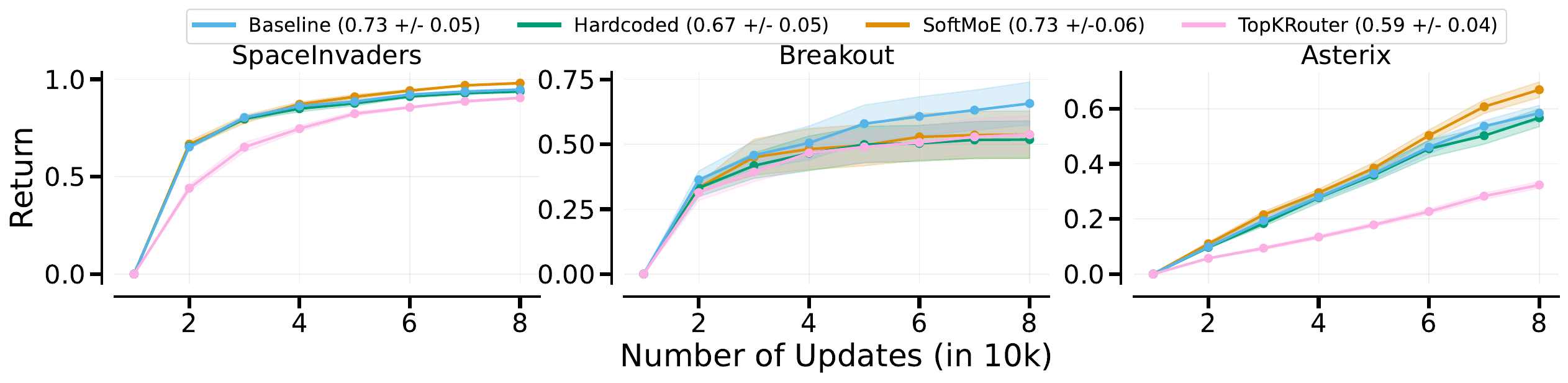}%
    \caption{\textbf{Left:} Aggregate Interquantile Mean \citep{agarwal2021deep} of scores. \textbf{Right} Comparison of different models and routers on the single environment setup. We report the mean of the normalised scores for 3 Atari games. All games run with 10 independent seeds, shaded areas representing the standard error. We normalise performance according to the single environment results reported in \citet{jesson2023relu}. BigMoE improves performance over the baseline, especially due to performance improvements in Asterix.}
    %\vspace{-5mm}}
    \label{fig:single_env_top}
\end{figure*}

\begin{table}[h]
\centering
\begin{tabular}{|l|c|c|c|c|}
\hline
Game & Big-SoftMoE & Baseline & Big-Hardcoded & Big-TopK \\
\hline
SI   & $0.98 \pm 0.01$ & $0.95 \pm 0.01$ & $0.94 \pm 0.01$ & $0.90 \pm 0.01$ \\
BO   & $0.54 \pm 0.09$ & $0.66 \pm 0.08$ & $0.52 \pm 0.07$ & $0.54 \pm 0.07$ \\
Ast  & $0.67 \pm 0.03$ & $0.58 \pm 0.03$ & $0.57 \pm 0.03$ & $0.32 \pm 0.01$ \\
\hline
Total & $\mathbf{0.73 \pm 0.06}$ & $\mathbf{0.73 \pm 0.05}$ & $\mathbf{0.67 \pm 0.05}$ & $0.59 \pm 0.04$ \\
\hline
\end{tabular}
\caption{\textbf{Single environment: Normalised Performance of algorithms across games with average total performance}. We normalise performance according to the single environment results reported in \citet{jesson2023relu}. We do not achieve the same performance as we use smaller networks due to computational limits.}
\label{tab:single_env}
\end{table}

\begin{figure*}[hbt!]
    \centering
    \includegraphics[width=\textwidth]{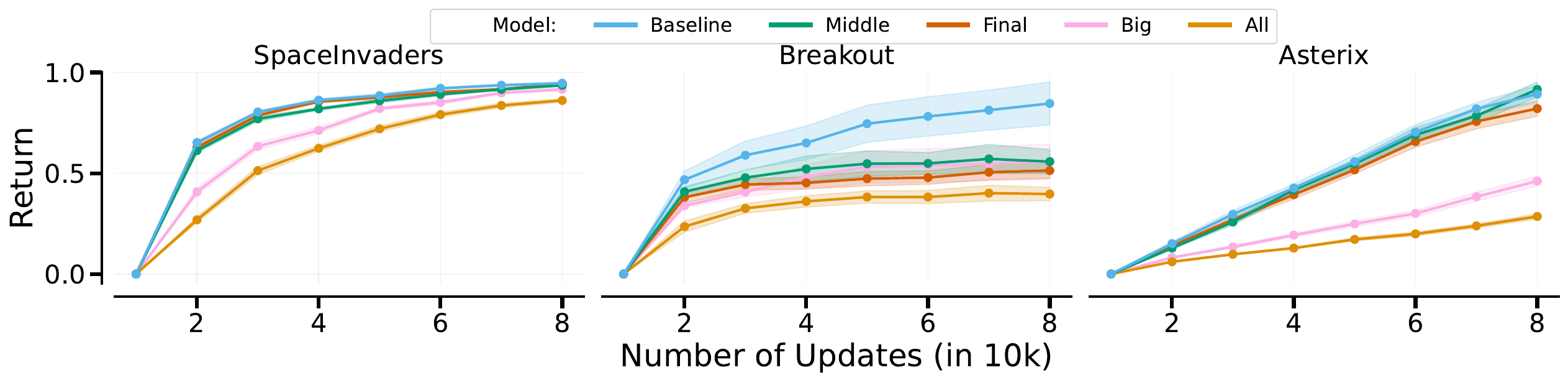}
    %single_env/topk/single_env_topk.png}
    \caption{\textbf{Single environment: All TopK Architectures}, all variants perform worse than the baseline.}
    \label{fig:single_env_topk}
\end{figure*}

\begin{table}[hbt!]
\centering
\begin{tabular}{|l|c|c|c|c|c|}
\hline
Game & \allmoe & \finalmoe & \middlemoe & Baseline & \bigmoe \\
\hline
SI   & $0.86 \pm 0.01$ & $0.94 \pm 0.01$ & $0.94 \pm 0.00$ & $0.95 \pm 0.01$ & $0.92 \pm 0.01$ \\
BO   & $0.31 \pm 0.03$ & $0.40 \pm 0.03$ & $0.43 \pm 0.05$ & $0.66 \pm 0.08$ & $0.43 \pm 0.07$ \\
Ast  & $0.19 \pm 0.01$ & $0.54 \pm 0.02$ & $0.60 \pm 0.02$ & $0.58 \pm 0.03$ & $0.30 \pm 0.02$ \\
\hline
Total & $0.45 \pm 0.02$ & $0.63 \pm 0.02$ & $\mathbf{0.66 \pm 0.03}$ & $\mathbf{0.73 \pm 0.05}$ & $0.55 \pm 0.04$ \\
\hline
\end{tabular}
\caption{\textbf{Single environment: All TopK Architectures}, all variants perform worse than the baseline.}
\label{tab:single_env_topk}
\end{table}

\begin{figure*}[hbt!]
    \centering
    \includegraphics[width=\textwidth]{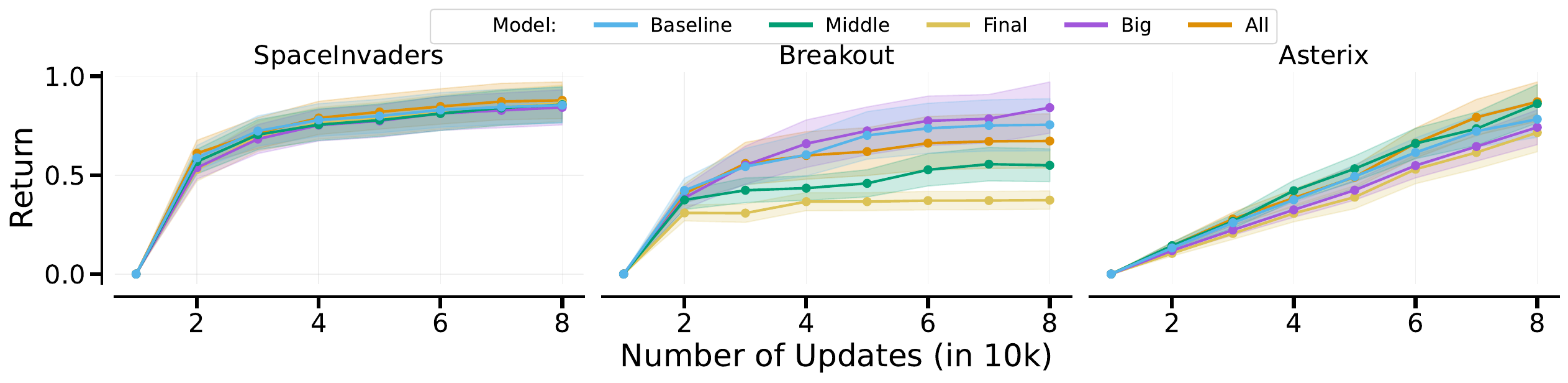}
    %single_env/softmoe/single_env_softmoe.png}
    \caption{\textbf{Single environment: All SoftMoE Architectures:} \bigmoe, \finalmoe, and \middlemoe all perform as well as the baseline.}
    \label{fig:single_env_softmoe}
\end{figure*}

\begin{table}[hbt!]
\centering
\begin{tabular}{|l|c|c|c|c|c|}
\hline
Game & \bigmoe & \allmoe & \finalmoe & Baseline & \middlemoe \\
\hline
SI   & $0.98 \pm 0.01$ & $0.96 \pm 0.00$ & $0.94 \pm 0.01$ & $0.95 \pm 0.01$ & $0.95 \pm 0.01$ \\
BO   & $0.54 \pm 0.09$ & $0.31 \pm 0.02$ & $0.67 \pm 0.08$ & $0.66 \pm 0.08$ & $0.44 \pm 0.05$ \\
Ast  & $0.67 \pm 0.03$ & $0.55 \pm 0.04$ & $0.56 \pm 0.03$ & $0.58 \pm 0.03$ & $0.65 \pm 0.02$ \\
\hline
Total & $\mathbf{0.73 \pm 0.06}$ & $0.61 \pm 0.03$ & $\mathbf{0.72 \pm 0.05}$ & $\mathbf{0.73 \pm 0.05}$ & $\mathbf{0.68 \pm 0.03}$ \\
\hline
\end{tabular}
\caption{\textbf{Single environment: All SoftMoE Architectures:} \bigmoe, \finalmoe, and \middlemoe all perform as well as the baseline.}
\label{tab:single_env_softmoe}
\end{table}

\begin{figure*}[hbt!]
    \centering
    \includegraphics[width=\textwidth]{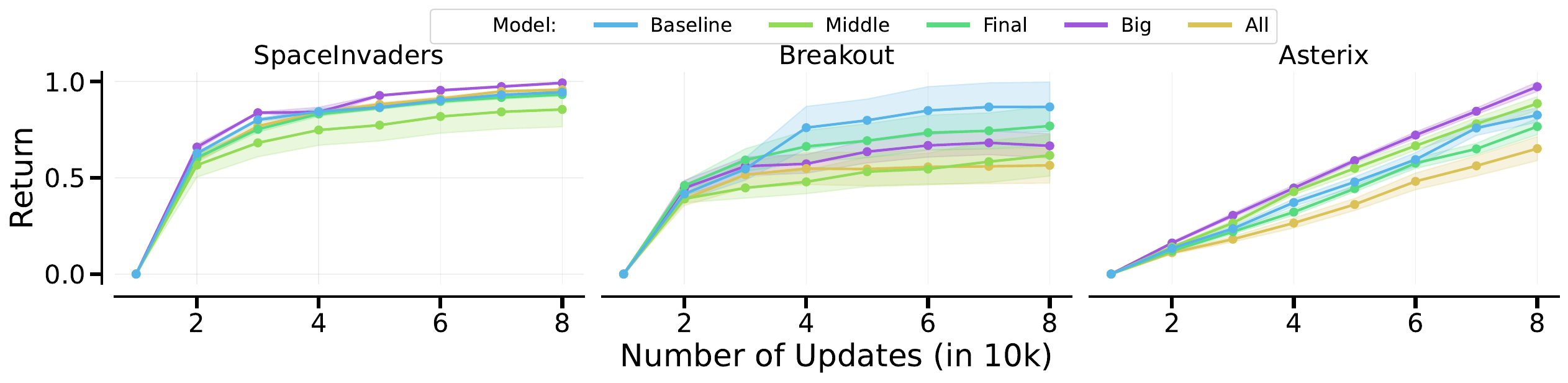}
    %single_env/softgradientmoe/single_env_softgradmoe.png}
    \caption{\textbf{Single environment: All SoftGradientMoE Architectures.}, adding gradient information does not change the conclusions of the softmoe architectures above.}
    \label{fig:single_env_softgradmoe}
\end{figure*}

\begin{table}[hbt!]
\centering
\begin{tabular}{|l|c|c|c|c|c|}
\hline
Game & Baseline & \middlemoe & \bigmoe & \finalmoe & \allmoe \\
\hline
SI   & $0.95 \pm 0.01$ & $0.86 \pm 0.09$ & $0.99 \pm 0.00$ & $0.93 \pm 0.01$ & $0.96 \pm 0.01$ \\
BO   & $0.60 \pm 0.09$ & $0.42 \pm 0.07$ & $0.46 \pm 0.04$ & $0.53 \pm 0.07$ & $0.39 \pm 0.06$ \\
Ast  & $0.59 \pm 0.03$ & $0.64 \pm 0.03$ & $0.70 \pm 0.02$ & $0.55 \pm 0.03$ & $0.47 \pm 0.04$ \\
\hline
Total & $\mathbf{0.71 \pm 0.05}$ & $\mathbf{0.64 \pm 0.07}$ & $\mathbf{0.72 \pm 0.03}$ & $\mathbf{0.67 \pm 0.05}$ & $0.60 \pm 0.04$ \\
\hline
\end{tabular}
\caption{\textbf{Single environment: All SoftGradientMoE Architectures}, adding gradient information does not change the conclusions of the softmoe architectures above.}
\label{tab:single_env_softgradmoe}
\end{table}

\begin{figure*}[hbt!]
    \centering
    \includegraphics[width=\textwidth]{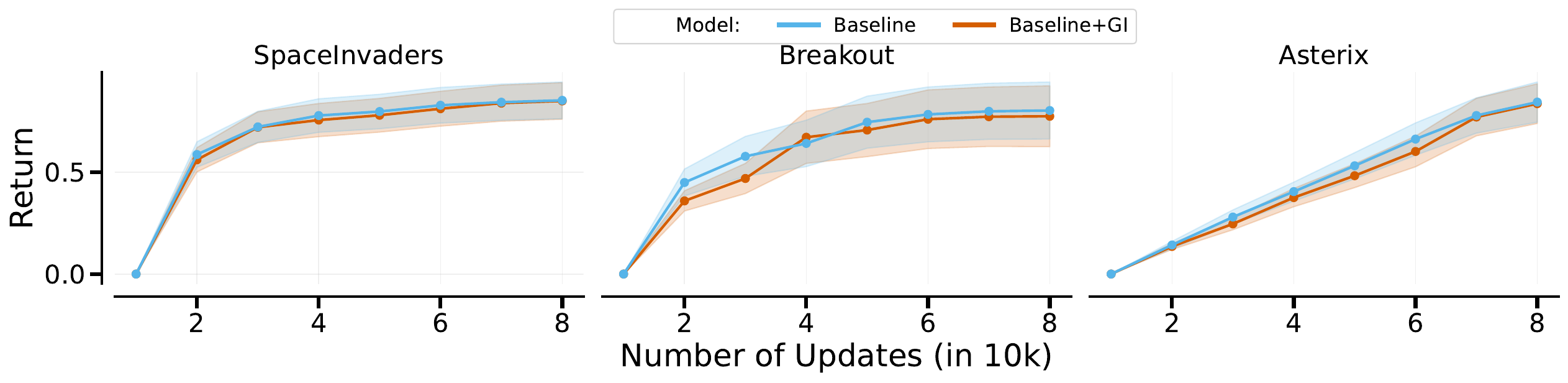}
    %single_env/vanilla_vs_vanilla/single_env_vanilla.png}
    \caption{\textbf{Single environment: Baseline vs. Baseline with Gradient Information}, adding gradient information to the input of the baseline does not affect its performance.}
    \label{fig:single_env_vanilla}
\end{figure*}

\begin{table}[hbt!]
\centering
\begin{tabular}{|l|c|c|}
\hline
Game & w/ Gradient Info & w/o Gradient Info \\
\hline
SI   & $0.95 \pm 0.01$ & $0.95 \pm 0.01$ \\
BO   & $0.66 \pm 0.08$ & $0.60 \pm 0.09$ \\
Ast  & $0.58 \pm 0.03$ & $0.59 \pm 0.03$ \\
\hline
Total & $\mathbf{0.73 \pm 0.05}$ & $\mathbf{0.71 \pm 0.05}$ \\
\hline
\end{tabular}
\caption{\textbf{Baseline vs. Baseline with Gradient Information:} Adding gradient information to the input of the baseline does not affect its performance.}
\label{tab:single_env_vanilla}
\end{table}

\clearpage
\section{Hyperparameters}
\label{app:hyperparams}
\begin{table}[ht]
\centering
\begin{tabular}{l|l}
\hline
\textbf{Hyperparameter} & \textbf{Value} \\
\hline
Number of Environments & 128 \\
Learning Rate & 9e-4 \\
Steps & 64 \\
Total Timesteps & 1e7 \\
Updates & total\_timesteps // num\_steps // num\_envs \\
Update Epochs & 10 \\
Minibatches & 8 \\
Minibatch Size & num\_envs * num\_steps // num\_minibatches \\
GAE-$\gamma$ & 0.99 \\
GAE-$\lambda$ & 0.7 \\
Clip $\epsilon$ & 0.2 \\
Entropy Coefficient & 0.01 \\
Value Function Coeffficient & 0.5 \\
Max Gradient Norm & 1.9 \\
Activation & relu \\
Environment & \{SpaceInvaders-MinAtar, Breakout-Minatar, Asterix-MinAtar\} \\
Anneal learning rate & True \\
\# Experts & 3 \\
Layer Size & 64 \\
Expert Hidden Size & 64 \\
Model & \{BigMoE, FinalLayer, AllLayers, MiddleLayer\} \\
MoE & \{SoftMoE, MoE, SoftGradientMoE\} \\
Expert & \{BigExpert, Expert\} \\
Router & \{TopKRouter, HardcodedRouter\} \\
Number of Selected Experts & 1 \\
Task ID & \{True, False\} \\
Actor MoE & \{True, False\} \\
Critic MoE & \{True, False\} \\
Gradient Buckets & 5 \\
Router Entropy & \{True, False\} \\
\hline
\end{tabular}
\caption{\textbf{Potential Hyperparameters configurations}. We did not run a grid search over all potential combinations but report meaningful selections.}
\label{tab:hyperparameters}
\end{table}

\end{document}